\pdfoutput=1

\documentclass[10pt,twocolumn,letterpaper,pagenumbers]{article}

\usepackage[]{cvpr} 
\usepackage[accsupp]{axessibility}  

\usepackage{graphicx}
\usepackage{amsmath}
\usepackage{amssymb}
\usepackage{booktabs}

\usepackage{bm}
\usepackage{bbm}
\usepackage{mathtools}
\usepackage{amsfonts}       
\usepackage{nicefrac}       
\usepackage{microtype}      
\usepackage{multirow}
\usepackage{adjustbox}
\usepackage{wrapfig}
\usepackage{placeins}

\renewcommand{\ie}{\textit{i}.\textit{e}., }
\renewcommand{\eg}{\textit{e}.\textit{g}., }

\newcommand{\dev}[1]{\footnotesize{$\pm$#1}}

\def\rvs{{\mathbf{s}}}

\def\rvx{{\mathbf{x}}}

\usepackage{color}
\definecolor{Red7}{rgb}{0.941, 0.243, 0.243}
\definecolor{Green7}{RGB}{55, 178, 77}
\definecolor{Tdgreen}{rgb}{0,0.4,0.7}
\usepackage{pifont}
\newcommand{\cmark}{\ding{51}}%
\newcommand{\xmark}{\ding{55}}%
\newcommand{\ck}{\color{Green7}{\cmark}}
\newcommand{\xk}{\color{Red7}{\xmark}}

\usepackage[pagebackref,breaklinks,colorlinks]{hyperref}
\hypersetup{citecolor=Tdgreen}

\usepackage[capitalize]{cleveref}
\crefname{section}{Sec.}{Secs.}
\Crefname{section}{Section}{Sections}
\Crefname{table}{Table}{Tables}
\crefname{table}{Tab.}{Tabs.}

\newcommand{\citet}{\cite}

\def\cvprPaperID{3976} 
\def\confName{CVPR}
\def\confYear{2023}

    \makeatletter
\def\@fnsymbol#1{\ensuremath{\ifcase#1\or \dagger\or *\or \ddagger\or
   \mathsection\or \mathparagraph\or \|\or **\or \dagger\dagger
   \or \ddagger\ddagger \else\@ctrerr\fi}}
    \makeatother

\begin{document}

\title{WinCLIP: Zero-/Few-Shot Anomaly Classification and Segmentation}


\author{
Jongheon Jeong$^{2*}$\thanks{Work done during an Amazon internship.}~~~~
Yang Zou$^{1}$\thanks{The authors contributed equally.}~~~~
Taewan Kim$^{1}$\\
Dongqing Zhang$^{1}$~~~~
Avinash Ravichandran$^{1}$\thanks{Work done as part of AWS AI Labs.}~~~~
Onkar Dabeer$^{1}$\\
$^{1}$~AWS AI Labs~~~~
$^{2}$~KAIST\\
}

\maketitle

\begin{abstract}
Visual anomaly classification and segmentation are vital for automating industrial quality inspection. The focus of prior research in the field has been on training custom models for each quality inspection task, which requires task-specific images and annotation. In this paper we move away from this regime, addressing zero-shot and few-normal-shot anomaly classification and segmentation. Recently CLIP, a vision-language model, has shown revolutionary generality with competitive zero-/few-shot performance in comparison to full-supervision. But CLIP falls short on anomaly classification and segmentation tasks. Hence, we propose window-based CLIP (WinCLIP) with (1) a compositional ensemble on state words and prompt templates and (2) efficient extraction and aggregation of window/patch/image-level features aligned with text. We also propose its few-normal-shot extension WinCLIP+, which uses complementary information from normal images. In MVTec-AD (and VisA), without further tuning, WinCLIP achieves \mbox{$91.8\%/85.1\%$ $(78.1\%/79.6\%)$} AUROC in zero-shot anomaly classification and segmentation while WinCLIP+ does \mbox{$93.1\%/95.2\%$ $(83.8\%/96.4\%)$} in 1-normal-shot, surpassing state-of-the-art by large margins.
\end{abstract}

\section{Introduction}
\label{sec:intro}

\begin{figure}[!ht]
  \centering
  \includegraphics[width=\linewidth]{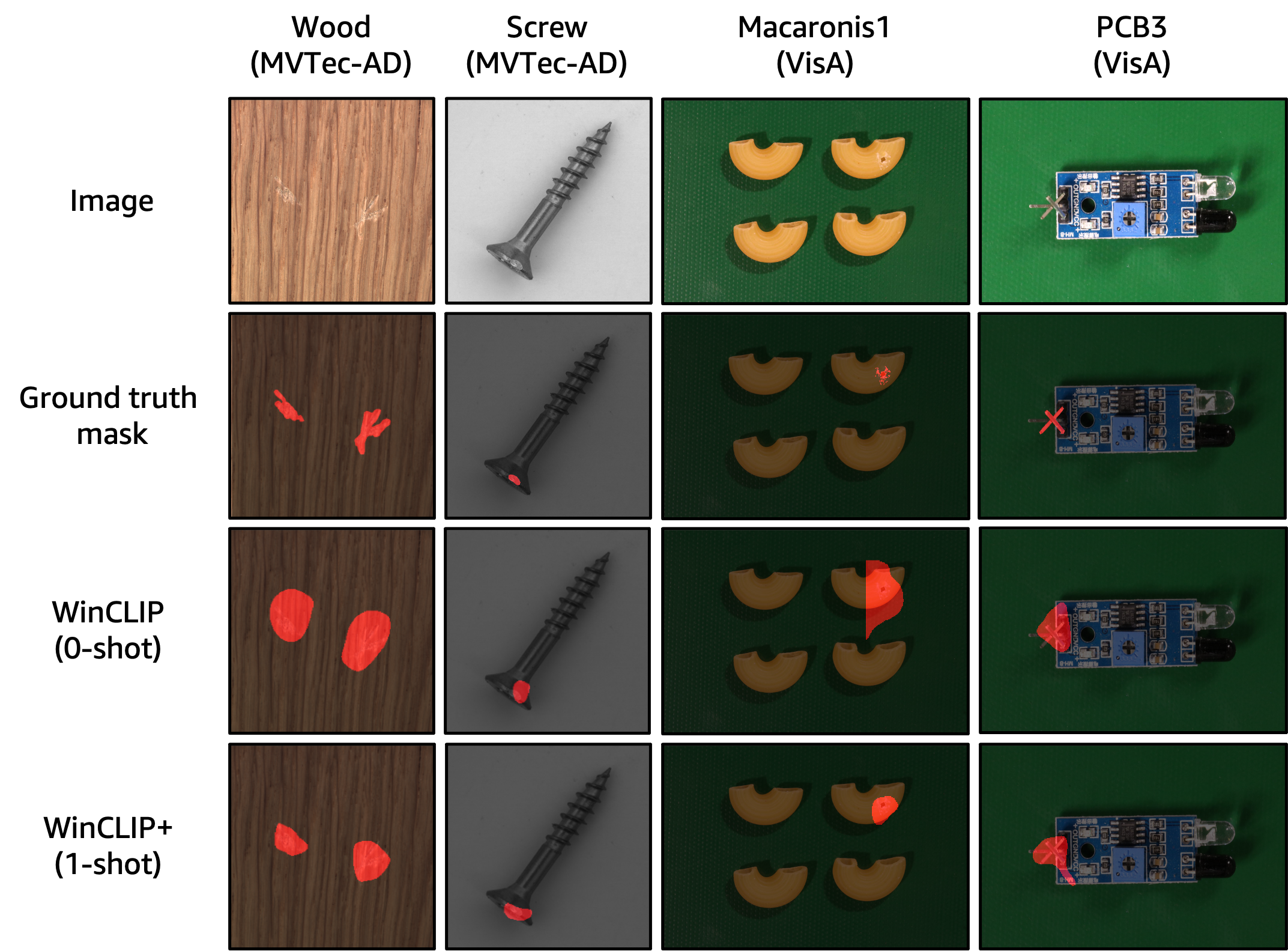}
\caption[Caption for LOF]{Language guided zero-/one-shot\protect\footnotemark anomaly segmentation from WinCLIP/WinCLIP+. Best viewed in color and zoom in.}
  \label{fig:vis_intro}
  \vspace{-0.2in}
\end{figure}

\footnotetext{\textit{few-shot} and \textit{few-normal-shot} are used interchangeably in our case.}
\begin{figure*}[!t]
  \centering
  \includegraphics[width=\linewidth]{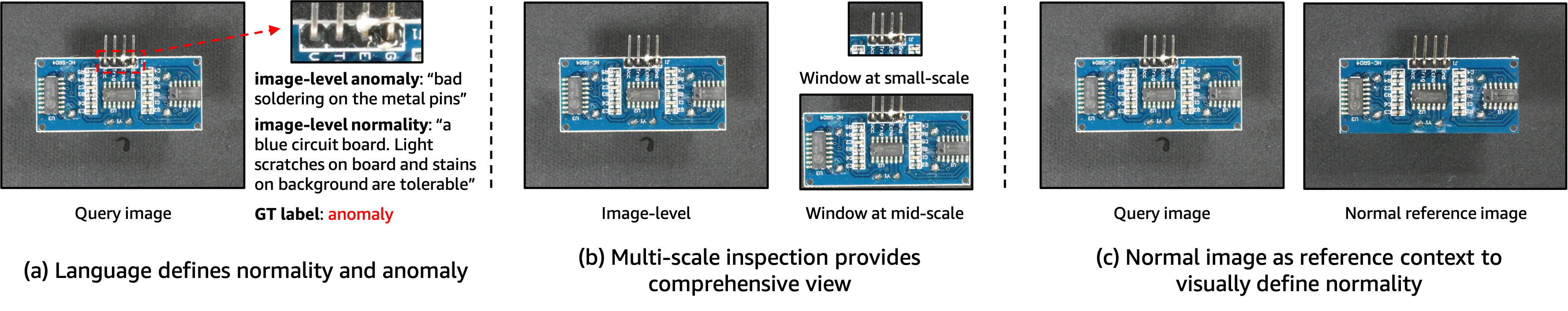}
  \caption{Motivation of language guided visual inspection. (a) Language helps describe and clarify normality and anomaly; (b) Aggregating multi-scale features helps identify local defects; (c) Normal images provide rich referencing content to visually define normality}
  \label{fig:principals}
  \vspace{-0.15in}
\end{figure*}

Visual anomaly classification (AC) and segmentation (AS) classify and localize defects in industrial manufacturing, respectively, predicting an image or a pixel as normal or anomalous. Visual inspection is a long-tail problem. The objects and their defects vary widely in color, texture, and size across a wide range of industrial domains, including aerospace, automobile, pharmaceutical, and electronics. These result in two main challenges in the field. 

First, defects are rare with wide range of variations, leading to a lack of representative anomaly samples in the training data. Consequently, existing works have mainly focused on one-class or unsupervised anomaly detection \cite{zou2022spot,li2021cutpaste,roth2022towards,defard2021padim,cohen2020sub,Ristea-CVPR-2022,zavrtanik2021draem,bergmann2022beyond}, which only requires normal images.  These methods typically fit a model to the normal images and treat any deviations from it as anomalous. When hundreds or thousands of normal images are available, many methods achieve high-accuracy on public benchmarks \cite{bergmann2019mvtec,defard2021padim,roth2022towards}. But in the few-normal-shot regime, there is still room to improve performance \cite{huang2022registration,zou2022spot,rudolph2021same,sheynin2021hierarchical}, particularly in comparison with the fully-supervised upper bound.

Second, prior work has focused on training a bespoke model for each visual inspection task, which is not scalable across the long-tail of tasks. This motivates our interest in zero-shot anomaly classification and segmentation. But many defects are defined with respect to a normal image. For example, a missing component on a circuit board is most easily defined with respect to a normal circuit board with all components present. For such cases, at least a few normal images are needed. So in addition to the zero-shot case, we also consider the case of few-normal-shot anomaly classification and segmentation. Since only few normal images are available, there is no segmentation supervision for localizing anomalies, making this a challenging problem across the long-tail of tasks.

Vision-language models \cite{radford2021learning,alayrac2022flamingo,jia2021scaling,schuhmann2022laionb} have shown promise in zero-shot classification tasks. Large-scale training with vision-language annotated pairs learns expressive representations that capture broad concepts. Without additional fine-tuning, text prompts can then be used to extract knowledge from such models for zero-/few-shot transfer to downstream tasks including image classification \cite{radford2021learning}, object detection \cite{gu2021open} and segmentation \cite{xu2022simple}. Since CLIP is one of the few open-source vision-language models, these works build on top of CLIP, benefiting from its generalization ability, and showing competitive low-shot performances in both seen and unseen objects compared to full supervision.


In this paper, we focus on zero-shot and few-normal-shot (1 to 4) regime, which has received limited attention \cite{rudolph2021same, sheynin2021hierarchical,huang2022registration}.
Our hypothesis is that language is perhaps even more important for zero-shot/few-normal-shot anomaly classification and segmentation. This hypothesis stems from multiple observations.
 First, ``normal'' and ``anomalous'' are states \cite{isola2015discovering} of an object that are context-dependent, and language helps clarify these states. For example, ``a hole in a cloth" may be a desirable or undesirable  depending upon whether distressed fashion or regular fashion clothes are being manufactured. Language can bring such context and specificity to the broad ``normal'' and ``anomalous'' states. Second, language can provide additional information to distinguish defects from acceptable deviations from normality. For example, in Figure~\ref{fig:principals}\textcolor{red}{(a)}, language provides information on the soldering defect, while minor scratches/stains on background are acceptable. In spite of these advantages, we are not aware of prior work leveraging vision-language models for anomaly classification and segmentation. In this work, with the pre-trained CLIP as a base model, we show and verify our hypothesis that language aids zero-/few-shot anomaly classification/segmentation.

 Since CLIP is one of the few open-source vision-language models, we build on top of it. Previously, CLIP-based methods have been applied for zero-shot classification \cite{radford2021learning}. CLIP can be applied in the same way to anomaly classification, using text prompts for ``normal'' and ``anomalous'' as classes. However, we find na\"ive prompts are not effective (see Table \ref{tab:ad_0shot}). So we improve the na\"ive baseline with a state-level word ensemble to better describe normal and anomalous states. Another challenge is that CLIP is trained to enforce cross-modal alignment only on the global embeddings of image and text. However, for anomaly segmentation we seek pixel-level classification and it is non-trivial to extract dense visual features aligned with language for zero-shot anomaly segmentation. Therefore, we propose a new \emph{Window-based CLIP} (WinCLIP), which extracts and aggregates the multi-scale features while ensuring vision-language alignment. The multiple scales used are illustrated in Figure~\ref{fig:principals}\textcolor{red}{(b)}. To leverage normal images available in the few-normal-shot setting, we introduce WinCLIP+, which aggregates complementary information from the language driven WinCLIP and visual cues from the normal reference images, such as the one shown in Figure~\ref{fig:principals}\textcolor{red}{(c)}. We emphasize that our zero-shot models do not require any tuning for individual cases, and the few-normal-only setup does not use any segmentation annotation, facilitating applicability across a broad range of visual inspection tasks. As a sample, Figure~\ref{fig:vis_intro} illustrates WinCLIP and WinCLIP+ qualitative results for a few cases. 

To summarize, our main contributions are:
\begin{itemize}
    \item We introduce a compositional prompt ensemble, which improves zero-shot anomaly classification over the na\"ive CLIP based zero-shot classification. 
    \item Using the pre-trained CLIP model, we propose WinCLIP, that  efficiently extract and aggregate multi-scale spatial features aligned with language for zero-shot anomaly segmentation. As far as we know, we are the first to explore language-guided zero-shot anomaly classification and segmentation. 
    \item We propose a simple reference association method, which is applied to multi-scale feature maps for image based few-shot anomaly segmentation. WinCLIP+ combines the language-guided and vision-only methods for few-normal-shot anomaly recognition.
    \item We show via extensive experiments on MVTec-AD and VisA benchmarks that our proposed methods WinCLIP/WinCLIP+ outperform the state-of-the-art methods in zero-/few-shot anomaly classification and segmentation with large margins.
\end{itemize}

\section{Related work}
\label{sec:related_work}

\noindent\textbf{Vision-language modeling. } 
Among recent successes of large pre-trained vision-language models (VLM) \cite{radford2021learning,jia2021scaling,alayrac2022flamingo}, CLIP \cite{radford2021learning} is the first to perform pre-training on web-scale image-text data, showing unprecedented generality: \eg its language-driven zero-shot inference, improved both effective robustness \cite{taori2020measuring} and perceptual alignment \cite{goh2021multimodal}.
Many following VLM works explored large-scale pre-training in different aspects, \eg scaling up data \cite{jia2021scaling}, efficient designs \cite{li2021align, alayrac2022flamingo, yang2022unified}, multi-tasks \cite{wang2022ofa,lu2022unified}, \textit{etc.} To democratize large-scale VLM for the usages in different domains, a billion-scale data LAION-5B \cite{schuhmann2022laionb}, a code base of OpenCLIP with pre-trained models \cite{ilharco_gabriel_2021_5143773} are open-sourced. Other works presented CLIP's promise in zero-/few-shot transfer to downstream tasks beyond classification \cite{tewel2022zerocap,gu2021open,xu2022simple,rombach2021highresolution}. 
Good prompt engineering and tuning can non-trivially benefit generalization performances \cite{radford2021learning,zhou2022learning}. Moreover, some other works \cite{zhou2022extract,rao2022denseclip,zhong2022regionclip} leverage the pre-trained CLIP for language guided detection and segmentation with promising performances. 

\vspace{0.05in}
\noindent\textbf{Anomaly classification and segmentation. } 
Due to the scarcity of anomalies, the major focus has been on one-class methods with many normal images \cite{li2021cutpaste,defard2021padim,cohen2020sub,yi2020patch,zavrtanik2021draem,yu2021fastflow}. While the MVTec-AD benchmark \cite{bergmann2019mvtec} is saturated by several works \cite{roth2022towards,yu2021fastflow,yang2022memseg}, their specific application is hindered due to their unscalable full-normal-shot setup. Recent works \cite{rudolph2021same,sheynin2021hierarchical} explored few-shot setups by leveraging augmentation to expand the small support set for better normality modeling. RegAD \cite{huang2022registration} further proposed a model-reusing by pre-training an object-agnostic registration network with diverse images to model normality for unseen object, given a few normal samples. Meanwhile, to close the gap between academical and industrial data, Visual Anomaly (VisA) \cite{zou2022spot} is introduced for a challenging benchmark over MVTec-AD. 
Additionally, Vision Transformer (ViT) have recently shown its potential in visual inspection \cite{mishra21-vt-adl,fort2021exploring}. 

\vspace{0.05in}
\noindent\textbf{State classification. } 
In some sense, anomaly classification is related to state classification \cite{isola2015discovering} that predicts if an object is normal or anomalous. While the major works in computer vision focus on object, scene, or material recognition \cite{russakovsky2015imagenet,hinton2012imagenet,xiao2010sun,sharan2013recognizing}, state classification aims to differentiate the fine-grained sub-object physical properties or attributes. Several datasets covering generic states/attributes (e.g. tall, crack, red, smooth) over diverse objects and scenes are introduced \cite{isola2015discovering,yu2014fine,hudson2019gqa,mancini2021open}. Some works \cite{welling2016semi,mancini2022learning,naeem2021learning} built graphs consisting of attributes and objects, of which relationship is learnt by graph neural networks \cite{zhang2019graph}. 

\section{Background}
\label{sec:preliminary}


\noindent\textbf{Anomaly classification and segmentation. }
Given an image $\rvx \in \mathcal{X}$, both anomaly classification and segmentation (ACS) aim to predict ``abnormality'' in $\rvx$.
Specifically, we consider anomaly classification (AC) as a binary classification $\mathcal{X} \rightarrow \{-, +\}$ where ``$+$'' indicates the presence of anomaly in image-level. And anomaly segmentation (AS) is its pixel-level extension to output the location of anomalies via $\mathcal{X} \rightarrow \{-, +\}^{h\times w}$ for a certain image with size $h \times w$. In practice, the tasks are often cast into problems of predicting anomaly scores. For example, anomaly classification typically models a mapping $\mathrm{ascore}: \mathcal{X}\rightarrow [0, 1]$
so that a binary classification can be performed by thresholding $\mathrm{ascore}(\rvx)$. 


Due to the lack of anomalous (or positive) samples in practice, the one-class scenario, where the training data $\mathcal{D}:=\{(x_i, -)\}_{i=1}^{K}$ consists of only normal (or negative) samples, has been widely used. 
In this paper, we follow the one-class protocol, particularly focusing on extreme cases of few-shot ($K=1$ to $4$) and the unexplored zero-shot setups for both AC and AS. And we assume an available list of task-specific texts tags, \eg for objects and relevant defects. 


\vspace{0.05in}
\noindent\textbf{Zero-shot classification with CLIP. }
\emph{Contrastive Language Image Pre-training} (CLIP) \cite{radford2021learning} is a large-scale pre-training method offering a joint vision-language representation.
Given million-scale image-text pairs $\{(x_t, s_t)\}_{t=1}^T$ from the web, CLIP trains an image encoder $f$ and a text encoder $g$ via {contrastive learning} \cite{zhang2020contrastive,chen2020simple} to maximize 
the correlation between $f(x_t)$ and $g(s_t)$ across $t$ in terms of cosine similarity $\langle f(\rvx), g(\rvs) \rangle$. 
Given an input $\rvx$ and a closed set of free-form texts $S=\{ s_1, \cdots, s_k \}$, CLIP can perform zero-shot classification via a $k$-way categorical distribution:
\begin{equation}\label{eq:clip_0shot}
    p(\rvs = s_i|\rvx; \rvs \in S) := \frac{\exp(\langle f(\rvx), g(s_i) \rangle / \tau) }{\sum_{s \in S} \exp(\langle f(\rvx), g(s) \rangle / \tau) },
\end{equation}
where $\tau > 0$ is the temperature hyperparameter. 

For a set of class words $C = \{c_1, \cdots, c_k\}$, it has shown that accompanying each label word $c\in C$ with a \emph{prompt template}, \eg ``\texttt{a photo of a [c]}'', improves accuracy over the case without templates. Moreover, an ensemble of prompt embeddings that aggregates multiple (80) templates \eg ``\texttt{a cropped photo of a [c]}'', can further boost the performance \cite{radford2021learning}. 
Overall, we are essentially ``retrieving'' the visual knowledge of CLIP through the language interface in appropriate manners. In this paper, we further explore how to extract the knowledge of CLIP in a way more suitable for anomaly recognition.

\section{WinCLIP and WinCLIP+}
\label{sec:method}


In this section, we first establish a novel binary zero-shot anomaly classification framework with a Compositional Prompt Ensemble to improve CLIP for anomaly classification (Section~\ref{sec:zeroshot}). Next, we propose a simple-yet-effective \emph{Window-based CLIP} (WinCLIP) for efficient zero-shot anomaly segmentation (Section~\ref{sec:winclip}). Lastly, we propose an extension \emph{WinCLIP+} to benefit from few normal reference images, while maintaining the complementary benefits of language-guided predictions (Section~\ref{sec:winclip+}).  



\subsection{Language-driven zero-shot AC}
\label{sec:zeroshot}



\noindent\textbf{Two-class design. }
We introduce a binary zero-shot anomaly classification framework \emph{CLIP-AC} by adapting CLIP with two class prompts \verb|[c]| - ``\verb|normal [o]|'' \textit{vs.}~``\verb|anomalous [o]|''. \verb|[o]| is an object-level label, \eg ``\verb|bottle|'' when available, or simply ``\verb|object|''. In addition, we also test a one-class design by only using the normal prompt $s_{-}:=\text{``\texttt{normal [o]}''}$ to define anomaly score as ``$-\langle f(\rvx), g(s_{-}) \rangle$''. We observe the simple two-class design from CLIP already yields a non-trial performance and outperforms one-class design significantly in experiments (Table~\ref{tab:ad_0shot}). This demonstrates (a) CLIP pre-trained by large web dataset provides a powerful representation with good alignment between text and images for anomaly tasks (b) specific definition about anomaly is necessary for good performance. 
\noindent\textbf{Compositional prompt ensemble (CPE). }
Unlike object-level classifiers, CLIP-AC performs classification between two \emph{states} of a given object, \ie either ``normal'' or ``anomalous'', which are subjective with various definitions depending on tasks. For example, ``missing transistor'' is ``anomalous'' for a circuit board while ``cracked'' is ``anomalous'' for wood. To better define the two abstract states of objects, we propose a Compositional Prompt Ensemble to generate all combinations of pre-defined lists of (a) \emph{state words} per label and (b) \emph{text templates}, rather than freely writing definitions. The state words include common states shared by most objects, \eg ``flawless'' for normality/``damaged'' for anomaly. Also we can optionally add task-specific state words given prior knowledge of defects, \eg ``bad soldering'' on PCB. Moreover, we curate a template list specifically for anomaly tasks \eg ``\texttt{a photo of a [c] for visual inspection}''. Check details on prompt engineering in supplementary. As in top-left of Figure ~\ref{fig:overall}, after getting all the combinations of states and templates, we compute the average of text embeddings per label to represent the normal and anomalous classes. Note that CPE is different from CLIP prompt ensemble that does not explain object labels (\eg ``cat'') and only augments templates selected by trial-and-error for object classification, including the ones unsuitable for anomaly tasks, \eg ``\verb|a cartoon [c]|''. Thus, the texts from CPE are more aligned with images in CLIP's joint embedding space for anomaly tasks. We denote the zero-shot scoring model with CPE as $\mathrm{ascore_0}: \mathbb{R}^d \rightarrow [0, 1]$ for an image embedding $f(\rvx)$.

\noindent\textbf{Remark. }
Our two-class design with CPE is a novel approach to define anomaly compared to standard one-class methods \cite{pmlr-v80-ruff18a,roth2022towards}. Anomaly detection is an ill-posed problem due to the open-ended nature. Previous methods model normality only by normal images regarding any deviation from normality as anomaly. Such solution is by nature hard to distinguish true anomalies from acceptable deviations from normality, \eg "scratch on circuit" vs. "tiny yet acceptable scratch". But language can define states in concrete words.

\begin{figure}[t]
  \includegraphics[width=0.48\textwidth]{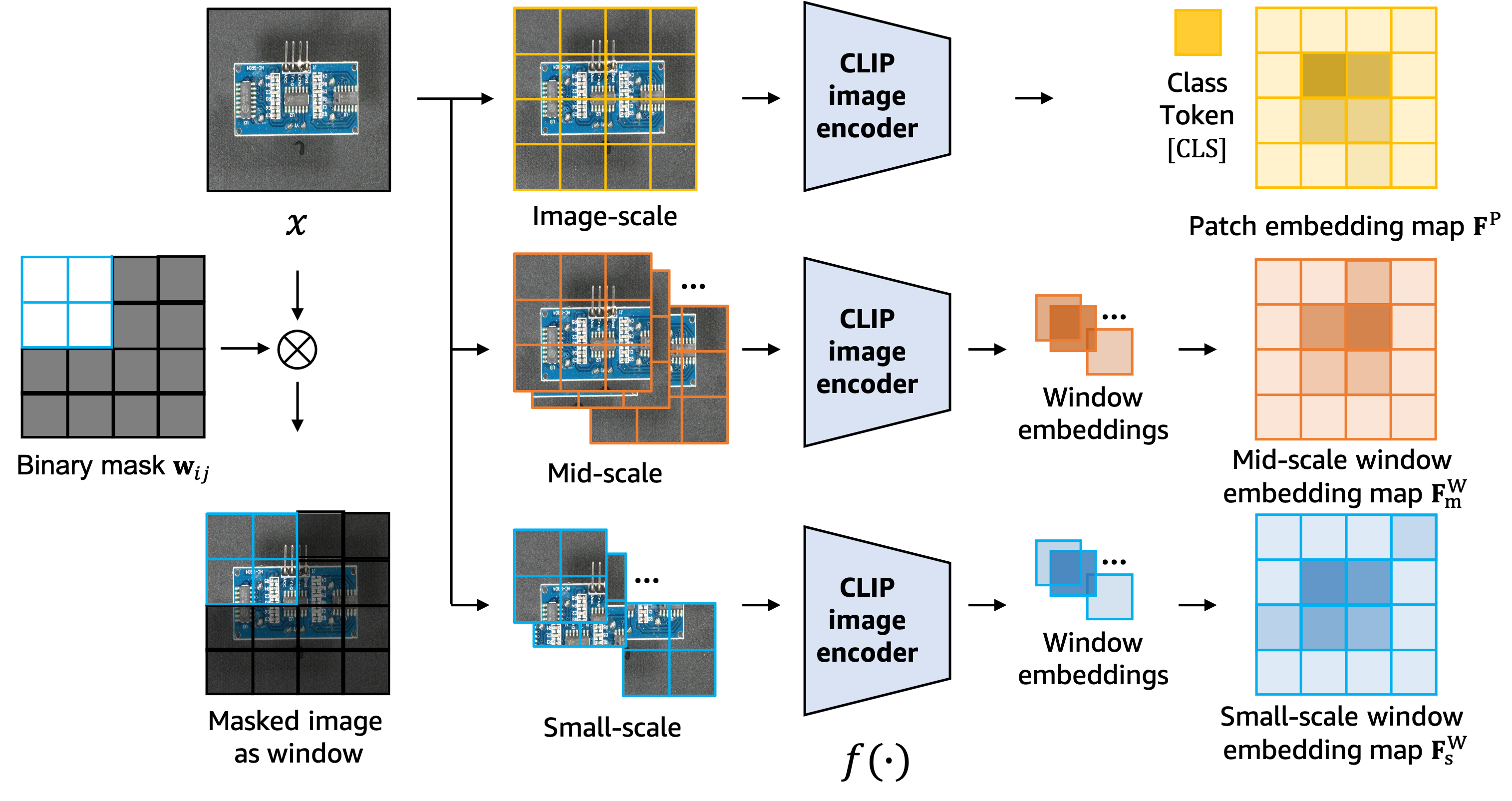}
  \caption{WinCLIP feature extraction in multiple scales of windows through CLIP image encoder, \eg ViT taking a sequence of (non-masked) patches as input. Window embeddings encode the global information (\eg from the class token) within each window.}
  \label{fig:featExtract}
  \vspace{-0.15in}
\end{figure}


\begin{figure*}[t]
  \centering
  \includegraphics[width=0.98\linewidth]{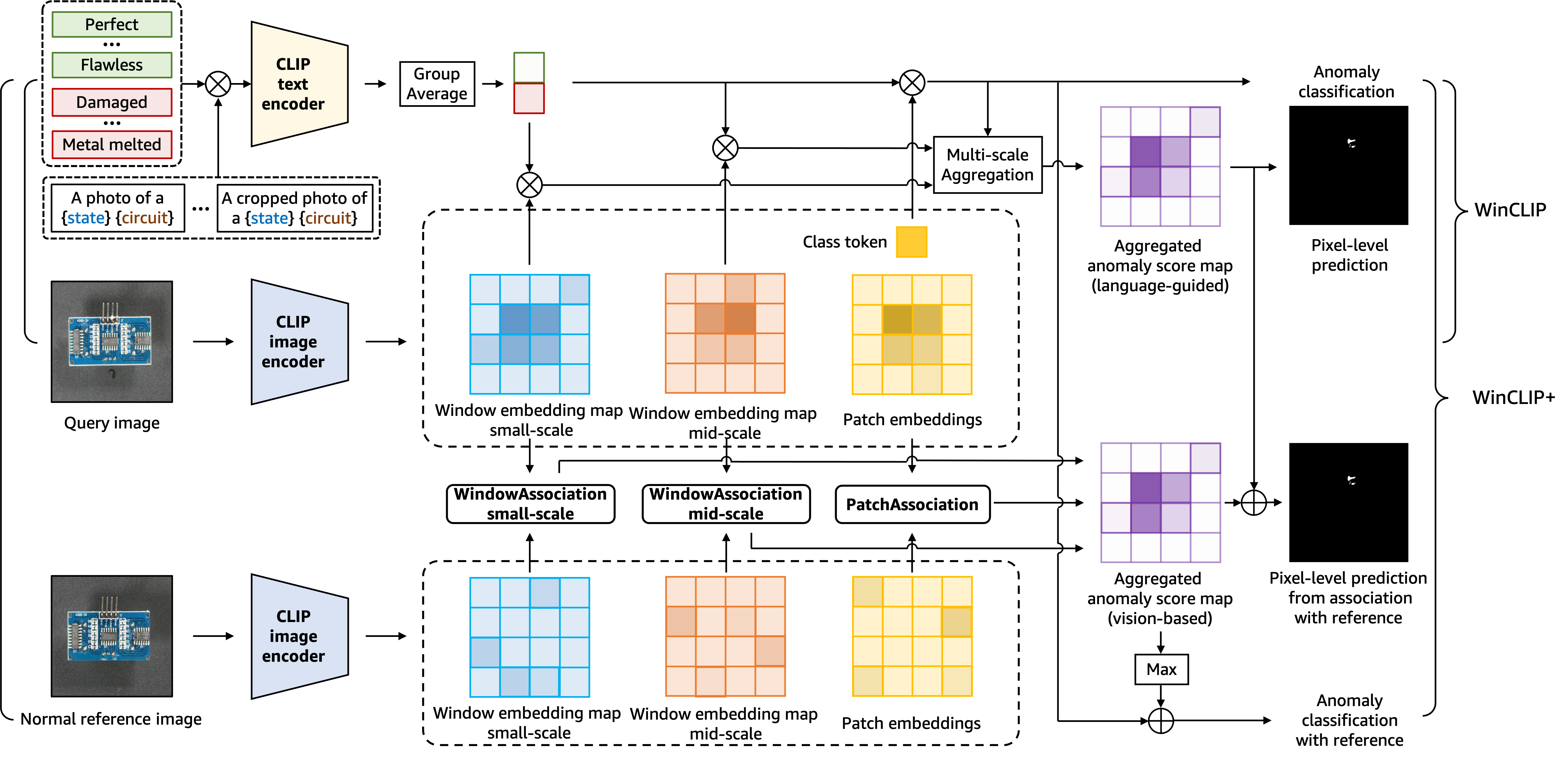}
  \vspace{-0.05in}
  \caption{Workflows of WinCLIP/WinCLIP+ (upper/entire pane). Various states and templates are composited and converted to two text embeddings as class prototypes via CLIP text encoder (Section \ref{sec:zeroshot}). The class prototypes are correlated with the multi-scale features from CLIP image encoder (Figure \ref{fig:featExtract}) for zero-shot AC/AS in WinCLIP. WinCLIP+ applies the reference association on patch, small-/mid-window (Patch/WindowAssociation) for vision-based anomaly score maps, which are aggregated for few-shot AS/AC with language-guided scores.}
  \label{fig:overall}
  \vspace{-0.15in}
\end{figure*}

\subsection{WinCLIP for zero-shot AS}
\label{sec:winclip}

Given the language guided anomaly scoring model from CPE, we propose Window-based CLIP (WinCLIP) for zero-shot anomaly segmentation to predict pixel-level anomalies. WinCLIP extracts dense visual features with good language alignment and local details for $\rvx$, followed by applying $\mathrm{ascore}_0$ spatially to obtain the anomaly segmentation map. Specifically, given an image $\rvx$ of resolution $h\times w$ and an image encoder $f$, \mbox{WinCLIP} obtains a map of $d$-dimensional feature map $\mathbf{F}^{\tt W} \in \mathbb{R}^{h\times w\times d}$ as follows:
\begin{enumerate}
    \item Generate a set of sliding windows $\{\mathbf{w}_{ij}\}_{ij}$, where each window $\mathbf{w}_{ij} \in \{0,1\}^{h \times w}$ is a binary mask that is active locally for a $k \times k$ kernel around $(i, j)$.
    \item Collect each output embedding $\mathbf{F}^{\tt W}_{ij}$, computed from the active area of $\rvx$ after applying each $\mathbf{w}_{ij}$, defined by:
    \begin{equation}\label{eq:winclip}
        \mathbf{F}^{\tt W}_{ij} := f(\rvx \odot \mathbf{w}_{ij}),
    \end{equation}
    where $\odot$ is the element-wise product (see Figure~\ref{fig:featExtract}).
\end{enumerate}
Figure~\ref{fig:featExtract} illustrates the dense feature extraction of WinCLIP with ViT while it is also applicable to CNN.


In addition, we also explore a natural dense representation candidate, \emph{penultimate feature map}, the last feature map before pooling. Specifically, for patch embedding map $\mathbf{F}^{\tt P}$ (other than the \emph{class token} \verb|[CLS]|) of ViT-based CLIP, top of Figure~\ref{fig:featExtract}, we apply $\mathrm{ascore}_0$ patch-wisely for segmentation. However, we observe that such patch-level features are not aligned with the language space, leading to a poor dense predictions (Table~\ref{tab:al_0shot}). We conjecture this is caused by those features have not been directly supervised with language signal in CLIP. Also these patch features have already aggregated the global context due to self-attention, hindering capturing local details for segmentation.

Compared to the penultimate features $\mathbf{F}^{\tt P}$, we remark dense features from WinCLIP is more aligned with language: \eg for ViT-based CLIP, all the features in $\mathbf{F}^{\tt W}$ are now from class tokens which are directly aligned to texts in CLIP pre-training. Also the features focus more on local details via sliding windows. Lastly, WinCLIP can be efficiently computed, especially with ViT architecture. Concretely, the computation of \eqref{eq:winclip} can directly benefit from just dropping all the masked patches before forwarding them, in a similar manner to masked autoencoder \cite{he2022masked}.

\vspace{0.05in}
\noindent\textbf{Harmonic aggregation of windows. }
For each local window, the zero-shot anomaly score $\mathbf{M}^{\tt W}_{0,ij}$ is similarity between the window feature $\mathbf{F}^{\tt W}_{ij}$ and text embeddings from compositional prompt ensemble. This score is distributed to every pixel of the local window. Then at each pixel, we aggregate multiple scores from all overlapping windows to improve segmentation by \emph{harmonic averaging} \eqref{eq:harmonic}, weighting more on scores towards normality prediction (zero value).
\begin{equation}\label{eq:harmonic}
    \bar{\mathbf{M}}^{\tt W}_{0,ij} := \left(\frac{1}{\sum_{u,v} {(\mathbf{w}_{uv})_{ij}}} \sum_{u,v} \frac{(\mathbf{w}_{uv})_{ij}}{\mathbf{M}^{\tt W}_{0,uv}} \right)^{-1}.
\end{equation}

\vspace{0.05in}
\noindent\textbf{Multi-scale aggregation. } 
The kernel size $k$ corresponds to the amount of surrounding context for each location in computing WinCLIP features \eqref{eq:winclip}. It controls the balance between local details and global information in segmentation. To capture defects of sizes ranging from small to large scale, we aggregate predictions from multi-scale features: \eg (a) small-scale ($2\times 2$ in patch scales of ViT; corresponds to $32\times 32$ in pixels), (b) mid-scale ($3\times 3$ in ViT; $48\times 48$), and (c) image-scale feature (ViT class token capturing image context due to self-attention). We also adopt harmonic averaging for aggregation. Figure~\ref{fig:featExtract} illustrates the features on each scale.

\subsection{WinCLIP+ with few-normal-shots}
\label{sec:winclip+}
For a comprehensive anomaly classification and segmentation, language guided zero-shot approach is not enough as certain defects can only be defined via visual reference rather than only text. For example, ``Metal-nut'' in MVTec-AD \cite{bergmann2019mvtec} has an anomaly type labeled as ``flipped upside-down'', which can only be identified relatively from a normal image. To define and recognize the anomalies more precisely, we propose an extension of WinCLIP, \emph{WinCLIP+}, by incorporating $K$ normal reference images $\mathcal{D}:=\{(x_i, -)\}_{i=1}^{K}$. WinCLIP+ combines the complementary prediction from both language-guided and visual based approachs for better anomaly classification and segmentation.


We first propose a \emph{reference association} as the key module to incorporate given reference images, which can simply store and retrieve the memory features $\mathbf{R}$ of $\mathcal{D}$ based on the cosine similarity. Given such module and the corresponding (\eg patch-level\footnote{Nevertheless, the module is generally applicable for other scales.}) features $\mathbf{F} \in \mathbb{R}^{h\times w\times d}$ extracted from a query image, a prediction $\mathbf{M} \in [0, 1]^{h\times w}$ for anomaly segmentation can be made by: 
\begin{equation}\label{eq:memory_al}
    \mathbf{M}_{ij} := \min_{r \in \mathbf{R}} \tfrac{1}{2} (1 - \langle \mathbf{F}_{ij}, r \rangle).
\end{equation}

Then we apply this association module at multiple scales of feature maps that are obtained from WinCLIP (see Figure~\ref{fig:overall} for the overall illustration). Specifically, given few-shot samples, we construct separate reference memories from three different features: (a) WinCLIP features at small-scale $\mathbf{F}^{\tt W}_{\tt s}$, (b) those at mid-scale $\mathbf{F}^{\tt W}_{\tt m}$, and also (c) from penultimate features $\mathbf{F}^{\tt P}$ with global context (\eg the patch tokens in ViT capturing image context due to self-attention). Even though $\mathbf{F}^{\tt P}$ is not aligned with language, it still useful to define normality and anomaly. 

As a result, WinCLIP+ gets three reference memories: $\mathbf{R}^{\tt W}_{\tt s}$, $\mathbf{R}^{\tt W}_{\tt m}$, and $\mathbf{R}^{\tt P}$. 
Then, we average their multi-scale predictions (\ref{eq:memory_al}) for anomaly segmentation for a given query,
\begin{equation}\label{eq:winplus_al}
    \mathbf{M}^{\tt W} := \tfrac{1}{3}(\mathbf{M}^{\tt P} + \mathbf{M}^{\tt W}_{\tt s} + \mathbf{M}^{\tt W}_{\tt m}),
\end{equation}
and then fusing with our language-guided prediction $\bar{\mathbf{M}}^{\tt W}_0$.

To perform anomaly classification, we combine the maximum value of $\mathbf{M}^{\tt W}$ and the WinCLIP zero-shot classification score.
The two scores have complementary information to collaborative with, specifically (a) one from the spatial features of few-shot references, and (b) the other one from the CLIP knowledge retrieved via language:
\begin{equation}\label{eq:winplus_ad}
    \mathrm{ascore}_{\tt W}(\rvx) := \tfrac{1}{2}\left(\mathrm{ascore}_0(f(\rvx)) + \max_{ij} \mathbf{M}^{\tt W}_{ij} \right).
\end{equation}


\section{Experiments}
\label{sec:experiment}

We perform an array of experiments to evaluate the performance of WinCLIP-based ACS under low-shot regimes, covering recent challenging benchmarks on industrial anomaly classification and segmentation that we are focusing on. 
We also conduct an extensive ablation study to validate the individual effectiveness of our proposed components. The detailed setups, \eg pre-processing, metrics, and other implementation details, are given in the supplementary. 


\begin{table*}[t]
    \centering
    \small
    \hfill
    \begin{minipage}{0.69\linewidth}
        \centering
        \begin{adjustbox}{width=\linewidth}
\begin{tabular}{clcccccc}
\toprule
\multicolumn{2}{l}{Anomaly Classification} & \multicolumn{3}{c}{MVTec-AD} & \multicolumn{3}{c}{VisA} \\
\cmidrule(r){3-5} \cmidrule(l){6-8}
Setup & Method & AUROC & AUPR  & $F_1$-max  & AUROC & AUPR  & $F_1$-max \\
\midrule
\multirow{3}[2]{*}{\textbf{0-shot}} & CLIP-AC \cite{radford2021learning} & 74.0\dev{0.0} & 89.1\dev{0.0} & 88.5\dev{0.0} & 59.3\dev{0.0} & 67.0\dev{0.0} & 74.4\dev{0.0} \\
      &  + Prompt ens. \cite{radford2021learning} & 74.1\dev{0.0} & 89.5\dev{0.0} & 87.8\dev{0.0} & 58.2\dev{0.0} & 66.4\dev{0.0} & 74.0\dev{0.0} \\
\cmidrule{2-8}      & \textbf{WinCLIP (ours)} & \textbf{91.8\dev{0.0}} & \textbf{96.5\dev{0.0}} & \textbf{92.9\dev{0.0}} & \textbf{78.1\dev{0.0}} & \textbf{81.2\dev{0.0}} & \textbf{79.0\dev{0.0}} \\
\midrule
\multirow{4}[2]{*}{1-shot} & SPADE \cite{cohen2020sub} & 81.0\dev{2.0} & 90.6\dev{0.8} & 90.3\dev{0.8} & 79.5\dev{4.0} & 82.0\dev{3.3} & 80.7\dev{1.9} \\
      & PaDiM \cite{defard2021padim} & 76.6\dev{3.1} & 88.1\dev{1.7} & 88.2\dev{1.1} & 62.8\dev{5.4} & 68.3\dev{4.0} & 75.3\dev{1.2} \\
      & PatchCore \cite{roth2022towards} & 83.4\dev{3.0} & 92.2\dev{1.5} & 90.5\dev{1.5} & 79.9\dev{2.9} & 82.8\dev{2.3} & 81.7\dev{1.6} \\
\cmidrule{2-8}      & \textbf{WinCLIP+ (ours)} & \textbf{93.1\dev{2.0}} & \textbf{96.5\dev{0.9}} & \textbf{93.7\dev{1.1}} & \textbf{83.8\dev{4.0}} & \textbf{85.1\dev{4.0}} & \textbf{83.1\dev{1.7}} \\
\midrule
\multirow{4}[2]{*}{2-shot} & SPADE \cite{cohen2020sub} & 82.9\dev{2.6} & 91.7\dev{1.2} & 91.1\dev{1.0} & 80.7\dev{5.0} & 82.3\dev{4.3} & 81.7\dev{2.5} \\
      & PaDiM \cite{defard2021padim} & 78.9\dev{3.1} & 89.3\dev{1.7} & 89.2\dev{1.1} & 67.4\dev{5.1} & 71.6\dev{3.8} & 75.7\dev{1.8} \\
      & PatchCore \cite{roth2022towards} & 86.3\dev{3.3} & 93.8\dev{1.7} & 92.0\dev{1.5} & 81.6\dev{4.0} & 84.8\dev{3.2} & 82.5\dev{1.8} \\
\cmidrule{2-8}      & \textbf{WinCLIP+ (ours)} & \textbf{94.4\dev{1.3}} & \textbf{97.0\dev{0.7}} & \textbf{94.4\dev{0.8}} & \textbf{84.6\dev{2.4}} & \textbf{85.8\dev{2.7}} & \textbf{83.0\dev{1.4}} \\
\midrule
\multirow{4}[2]{*}{4-shot} & SPADE \cite{cohen2020sub} & 84.8\dev{2.5} & 92.5\dev{1.2} & 91.5\dev{0.9} & 81.7\dev{3.4} & 83.4\dev{2.7} & 82.1\dev{2.1} \\
      & PaDiM \cite{defard2021padim} & 80.4\dev{2.5} & 90.5\dev{1.6} & 90.2\dev{1.2} & 72.8\dev{2.9} & 75.6\dev{2.2} & 78.0\dev{1.2} \\
      & PatchCore \cite{roth2022towards} & 88.8\dev{2.6} & 94.5\dev{1.5} & 92.6\dev{1.6} & 85.3\dev{2.1} & 87.5\dev{2.1} & 84.3\dev{1.3} \\
\cmidrule{2-8}      & \textbf{WinCLIP+ (ours)} & \textbf{95.2\dev{1.3}} & \textbf{97.3\dev{0.6}} & \textbf{94.7\dev{0.8}} & \textbf{87.3\dev{1.8}} & \textbf{88.8\dev{1.8}} & \textbf{84.2\dev{1.6}} \\
\bottomrule
\end{tabular}%

        \end{adjustbox}
        \vspace{-0.05in}
        \caption{Comparison of anomaly classification (AC) performance on MVTec-AD and VisA benchmarks. We report the mean and standard deviation over 5 random seeds for each measurement. Bold indicates the best performance.}\label{tab:ad}
    \end{minipage}
    \hfill
    \begin{minipage}{0.27\linewidth}
        \centering
        \begin{adjustbox}{width=\linewidth}
\begin{tabular}{lccc}
\toprule
Methods & Setup & AC    & AS \\
\midrule
WinCLIP (ours) & 0-shot & 91.8  & 85.1 \\
WinCLIP+ (ours) & 1-shot & 93.1  & 95.2 \\
WinCLIP+ (ours) & 4-shot & 95.2  & 96.2 \\
\midrule
DifferNet \cite{rudolph2021same}  & 16-shot & 87.3  & - \\
TDG \cite{sheynin2021hierarchical}  & 10-shot & 78.0  & - \\
RegAD-L \cite{huang2022registration} & 2-shot & 81.5  & 93.3 \\
RegAD \cite{huang2022registration} & 4 + agg. & 88.2  & 95.8 \\
\midrule
MKD \cite{salehi2021multiresolution} & full-shot & 87.7  & 90.7 \\
P-SVDD \cite{yi2020patch} & full-shot & 92.1  & 95.7 \\
CutPaste \cite{li2021cutpaste} & full-shot & 95.2  & 96.0 \\
PatchCore \cite{roth2022towards} & full-shot & 99.6  & 98.2 \\
\bottomrule
\end{tabular}%

        \end{adjustbox}
        \vspace{-0.1in}
        \caption{Comparison with existing many-shot ACS methods in AUROC (or pixel-) on MVTec-AD.}
        \label{tab:mvtec_fs}
        \vspace{0.05in}
        \begin{adjustbox}{width=\linewidth}
\begin{tabular}{l|ccc}
\toprule
Method & AUROC & AUPR  & $F_1$-max \\
\midrule
(a) One-class & 34.2  & 68.9  & 83.5 \\
\midrule
Two-class & 74.0  & 89.1  & 88.5 \\
(b) + State ens. & 89.8  & 95.6  & 92.2 \\
(c) + Prompt ens. & 90.8  & 96.1  & 92.5 \\
(d) + Multi-crop & 91.8  & 96.5  & 92.9 \\
\bottomrule
\end{tabular}%

        \end{adjustbox}
        \vspace{-0.1in}
        \caption{Comparison of AC performance on MVTec-AD across WinCLIP ablations in AC (Section~\ref{sec:zeroshot}).}
        \label{tab:ad_0shot}
    \end{minipage}
    \hfill
    \vspace{-0.2in}
\end{table*}

\vspace{0.05in}
\noindent\textbf{Datasets. }
Our experiments are based on MVTec-AD \cite{bergmann2019mvtec} and VisA \cite{zou2022spot} datasets. Both benchmarks have diverse subsets of different objects, \eg capsules, circuit boards. They contain high-resolution images (\eg $700^2$-$1024^2$ for MVTec-AD, and roughly $1.5\text{K} \times 1\text{K}$ for VisA) of common objects with the full pixel-level annotations.

\vspace{0.05in}
\noindent\textbf{Evaluation metrics. }
For classification, we report (a) \emph{Area Under the Receiver Operating Characteristic} (AUROC) following the literature \cite{yi2020patch,defard2021padim,roth2022towards}, as well as (b) \emph{Area Under the Precision-Recall curve} (AUPR) and (c) \emph{$F_1$-score at optimal threshold} ($F_1$-max) for a clearer view against potential data imbalance \citet{zou2022spot}). For segmentation, we report (a) \emph{pixel-wise AUROC} (pAUROC) and (b) \emph{Per-Region Overlap} (PRO) \cite{bergmann2020uninformed} scores \cite{defard2021padim,li2021cutpaste}, and (c) \emph{(pixel-wise) $F_1$-max} in a similar manner to the anomaly classification evaluation. 

\vspace{0.05in}
\noindent\textbf{Implementation details. }
We adopt the CLIP implementation of OpenCLIP\footnote{\url{https://github.com/mlfoundations/open_clip}} and its public pre-trained models in our experiments: namely, we use the LAION-400M \cite{schuhmann2021laion} based CLIP with {ViT-B/16+} \cite{ilharco_gabriel_2021_5143773} unless otherwise noted. 
We apply WinCLIP with stride 1 on ViT patch embeddings, which is equivalent to stride 16 in pixel-level in case of ViT-B/16+. 




\subsection{Zero-/few-shot anomaly classification}
\label{sec:exp_ad}

In Table~\ref{tab:ad} we compare zero-shot and few-normal-shot anomaly classification results with prior works.

For zero-shot setup, we compare WinCLIP with two prior models: CLIP-AC (first row of Table~\ref{tab:ad}), which is the original 
CLIP zero-shot classification \cite{radford2021learning} with labels of the form $\{\text{``\texttt{normal [c]}''}, \text{``\texttt{anomalous [c]}''}\}$, and CLIP-AC with the prompt ensemble (second row in Table~\ref{tab:ad}) from \cite{radford2021learning} engineered for ImageNet \cite{krizhevsky2017imagenet}. We see that WinCLIP significantly improves over using these na\"ive adaptations of CLIP on both MVTec-AD and VisA. Section~\ref{sec:ablation} presents ablation study on a break-down of this gain.

For the few-normal-shot setup, we see the same trend: WinCLIP+ outperforms prior works by a wide margin across all metrics on both benchmarks. In particular, we improve upon the state-of-the-art PatchCore \cite{roth2022towards} by $9.7\%$ on 1-shot MVTec-AD and by $5.3\%$ on 1-shot VisA. On MVTec-AD, we note that zero-shot WinCLIP outperforms the few-shot versions of prior works. Furthermore, WinCLIP+ 1/2/4-shot performance is better than WinCLIP 0-shot performance, highlighting the additional value of reference normal images. 


\subsection{Zero-/few-shot anomaly segmentation}
\label{sec:exp_al}

In Table~\ref{tab:al} we compare zero-shot and few-normal-shot anomaly segmentation results with prior works. While there are no prior works on zero-shot anomaly segmentation, we adapt two methods developed for other problems to our setup. First, Trans-MM \cite{chefer2021generic} is a recent model interpretation method applicable to Transformers that provides a pixel-level mask. Second, MaskCLIP \cite{zhou2022extract} is a  general semantic segmentation model based on CLIP. We see that WinCLIP outperforms both methods by a wide margin on both MVTec-AD and VisA, highlighting that generic adaptations of CLIP do not perform as well as WinCLIP. 

\begin{figure*}[t]
    \centering
    \hspace*{\fill}
    \begin{subfigure}{0.40\linewidth}
        \includegraphics[width=\linewidth]{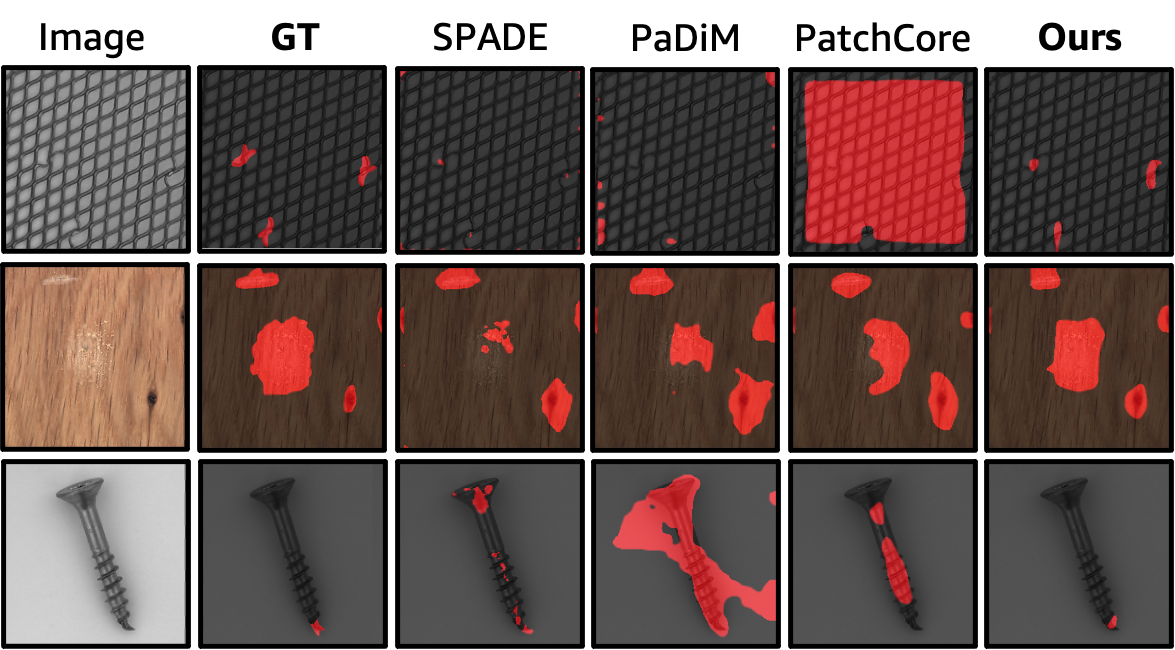}
        \caption{MVTec-AD (1-shot)}\label{fig:qual_mvtec}
    \end{subfigure}
    \hspace*{\fill}
    \begin{subfigure}{0.56\linewidth}
        \includegraphics[width=\linewidth]{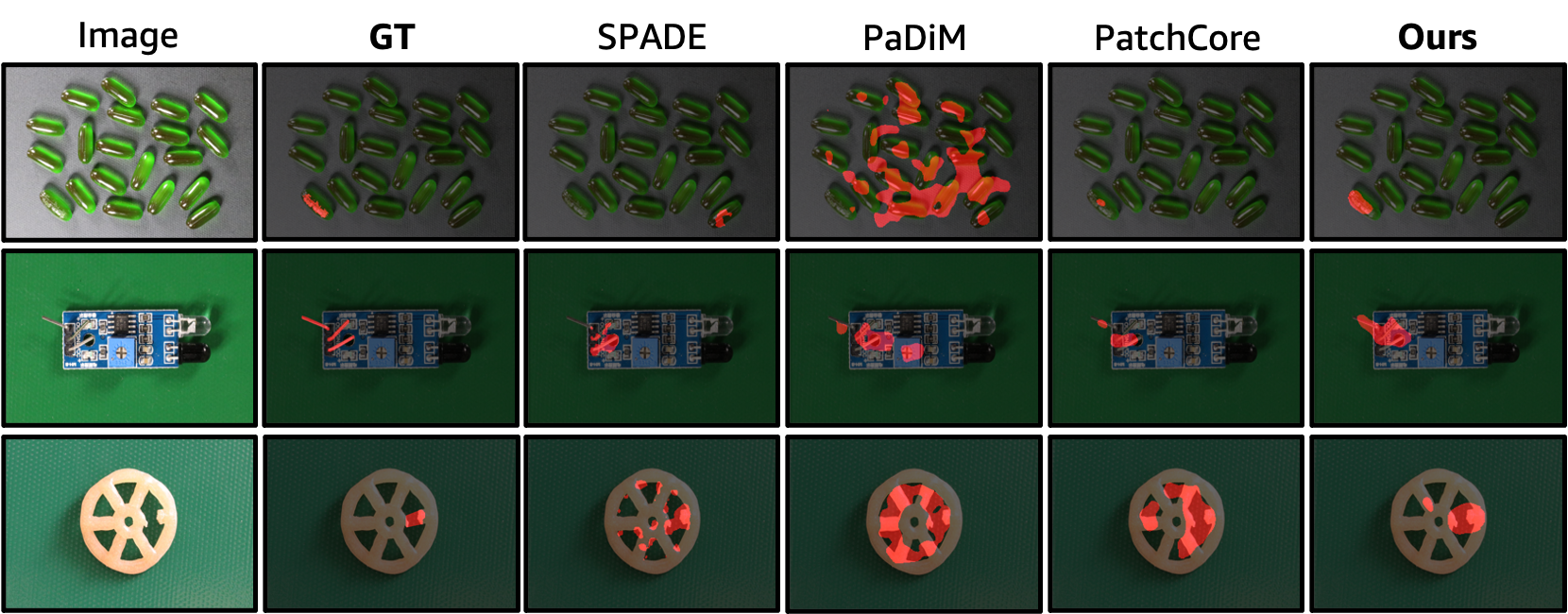}
        \caption{VisA (1-shot)}\label{fig:qual_visa}
    \end{subfigure}
    \hspace*{\fill}
    \vspace{-0.07in}
    \caption{Qualitative comparison of 1-shot anomaly segmentation results on MVTec-AD and VisA benchmarks.}
    \label{fig:qualitative}
    \vspace{-0.07in}
\end{figure*}

For the few-normal-shot setup, we compare with three prior works, which are designed specifically for anomaly localization. We see that WinCLIP+ again outperforms these prior methods across all metrics on both benchmarks, showing the additional value provided by language prompts. In Figure~\ref{fig:qualitative}, we show qualitative results for a number of objects and defects. We see that in all cases, 1-shot WinCLIP+ provides a mask that is more concentrated on the ground truth compared to prior works. We also see that 1/2/4-normal-shot WinCLIP+ is better than 0-shot WinCLIP, demonstrating the complementary benefits of language driven prediction and visual only based model based on reference normal images. 


\subsection{Comparison with many-shot methods}
In Table~\ref{tab:mvtec_fs} we compare our zero-/few-shot results with full-shot results of several prior works on MVTec-AD. Our 4-shot WinCLIP+ is competitive with CutPaste \cite{li2021cutpaste}, a recent method that utilizes the \textit{full-shot} samples for model tuning. 
Also, our 0-shot WinCLIP outperforms recent few-shot methods in AC, such as DifferNet \cite{rudolph2021same} and TDG \cite{sheynin2021hierarchical}, even compared to their results with more than 10-shots.
Recently, a new setup of aggregated few-shot is proposed \cite{huang2022registration}, where one is free to use all the training samples but for the target class which is restricted to $k$-shot. Our 4-shot WinCLIP+ outperforms RegAD's aggregated 4-shot \cite{huang2022registration} performance.

\begin{table*}[t]
    \centering
    \small
    \hfill
    \begin{minipage}{0.64\linewidth}
        \centering
        \begin{adjustbox}{width=\linewidth}
\begin{tabular}{clcccccc}
\toprule
\multicolumn{2}{l}{Anomaly Segmentation} & \multicolumn{3}{c}{MVTec-AD} & \multicolumn{3}{c}{VisA} \\
\cmidrule(r){3-5} \cmidrule(l){6-8}
Setup & Method & pAUROC & PRO   & $F_1$-max & pAUROC & PRO   & $F_1$-max \\
\midrule
\multirow{3}[2]{*}{\textbf{0-shot}} & Trans-MM  \cite{chefer2021generic} & 57.5\dev{0.0} & 21.9\dev{0.0} & 12.1\dev{0.0} & 49.4\dev{0.0} & 10.2\dev{0.0} & 3.1\dev{0.0} \\
      & MaskCLIP \cite{zhou2022extract} & 63.7\dev{0.0} & 40.5\dev{0.0} & 18.5\dev{0.0} & 60.9\dev{0.0} & 27.3\dev{0.0} & 7.3\dev{0.0} \\
\cmidrule{2-8}      & \textbf{WinCLIP (ours)} & \textbf{85.1\dev{0.0}} & \textbf{64.6\dev{0.0}} & \textbf{31.7\dev{0.0}} & \textbf{79.6\dev{0.0}} & \textbf{56.8\dev{0.0}} & \textbf{14.8\dev{0.0}} \\
\midrule
\multirow{4}[2]{*}{1-shot} & SPADE \cite{cohen2020sub} & 91.2\dev{0.4} & 83.9\dev{0.7} & 42.4\dev{1.0} & 95.6\dev{0.4} & 84.1\dev{1.6} & 35.5\dev{2.2} \\
      & PaDiM \cite{defard2021padim} & 89.3\dev{0.9} & 73.3\dev{2.0} & 40.2\dev{2.1} & 89.9\dev{0.8} & 64.3\dev{2.4} & 17.4\dev{1.7} \\
      & PatchCore \cite{roth2022towards} & 92.0\dev{1.0} & 79.7\dev{2.0} & 50.4\dev{2.1} & 95.4\dev{0.6} & 80.5\dev{2.5} & 38.0\dev{1.9} \\
\cmidrule{2-8}      & \textbf{WinCLIP+ (ours)} & \textbf{95.2\dev{0.5}} & \textbf{87.1\dev{1.2}} & \textbf{55.9\dev{2.7}} & \textbf{96.4\dev{0.4}} & \textbf{85.1\dev{2.1}} & \textbf{41.3\dev{2.3}} \\
\midrule
\multirow{4}[2]{*}{2-shot} & SPADE \cite{cohen2020sub} & 92.0\dev{0.3} & 85.7\dev{0.7} & 44.5\dev{1.0} & 96.2\dev{0.4} & 85.7\dev{1.1} & 40.5\dev{3.7} \\
      & PaDiM \cite{defard2021padim} & 91.3\dev{0.7} & 78.2\dev{1.8} & 43.7\dev{1.5} & 92.0\dev{0.7} & 70.1\dev{2.6} & 21.1\dev{2.4} \\
      & PatchCore \cite{roth2022towards} & 93.3\dev{0.6} & 82.3\dev{1.3} & 53.0\dev{1.7} & 96.1\dev{0.5} & 82.6\dev{2.3} & 41.0\dev{3.9} \\
\cmidrule{2-8}      & \textbf{WinCLIP+ (ours)} & \textbf{96.0\dev{0.3}} & \textbf{88.4\dev{0.9}} & \textbf{58.4\dev{1.7}} & \textbf{96.8\dev{0.3}} & \textbf{86.2\dev{1.4}} & \textbf{43.5\dev{3.3}} \\
\midrule
\multirow{4}[2]{*}{4-shot} & SPADE \cite{cohen2020sub} & 92.7\dev{0.3} & 87.0\dev{0.5} & 46.2\dev{1.3} & 96.6\dev{0.3} & 87.3\dev{0.8} & 43.6\dev{3.6} \\
      & PaDiM \cite{defard2021padim} & 92.6\dev{0.7} & 81.3\dev{1.9} & 46.1\dev{1.8} & 93.2\dev{0.5} & 72.6\dev{1.9} & 24.6\dev{1.8} \\
      & PatchCore \cite{roth2022towards} & 94.3\dev{0.5} & 84.3\dev{1.6} & 55.0\dev{1.9} & 96.8\dev{0.3} & 84.9\dev{1.4} & 43.9\dev{3.1} \\
\cmidrule{2-8}      & \textbf{WinCLIP+ (ours)} & \textbf{96.2\dev{0.3}} & \textbf{89.0\dev{0.8}} & \textbf{59.5\dev{1.8}} & \textbf{97.2\dev{0.2}} & \textbf{87.6\dev{0.9}} & \textbf{47.0\dev{3.0}} \\
\bottomrule
\end{tabular}%

        \end{adjustbox}
        \vspace{-0.06in}
        \caption{Comparison of anomaly segmentation (AS) performance on MVTec-AD and VisA benchmarks. We report the mean and standard deviation over 5 random seeds for each measurement. Bold indicates the best performance.}\label{tab:al}
    \end{minipage}
    \hfill
    \begin{minipage}{0.31\linewidth}
        \centering
        \begin{adjustbox}{width=0.98\linewidth}

\begin{tabular}{cccccc}
\toprule
\multicolumn{2}{c}{WinCLIP+ (AC)} & \multicolumn{4}{c}{\# shots (AUROC)} \\
\cmidrule(r){1-2} \cmidrule(l){3-6}
$\max \mathbf{M}^{\tt W}$ & \phantom{x}$\mathrm{ascore}_0$\phantom{x} & 1     & 2     & 4     & 8 \\
\cmidrule(r){1-2} \cmidrule(l){3-6}
\ck & \xk & 87.9  & 91.0  & \underline{92.6}  & \underline{94.5} \\
\xk & \ck & \underline{91.8}  & \underline{91.8}  & 91.8  & 91.8 \\
\cmidrule(r){1-2} \cmidrule(l){3-6}
\ck & \ck & \textbf{93.1}  & \textbf{94.4}  & \textbf{95.2}  & \textbf{96.3} \\
\bottomrule
\end{tabular}%

        \end{adjustbox}
        \vspace{-0.05in}
        \caption{$k$-shot AC ablations: MVTec-AD. Bold/underline indicate the best/runner-up.}\label{tab:winclipp_ad}
        \vspace{0.1in}
        \begin{adjustbox}{width=0.98\linewidth}
\begin{tabular}{ccccccc}
\toprule
\multicolumn{3}{c}{WinCLIP+ (AS)} & \multicolumn{4}{c}{\# shots (pAUROC)} \\
\cmidrule(r){1-3} \cmidrule(l){4-7}
$\mathbf{M}^{\tt P}$ & $\mathbf{M}^{\tt W}_{\tt m}$ & $\mathbf{M}^{\tt W}_{\tt s}$ & 1     & 2     & 4     & 8 \\
\cmidrule(r){1-3} \cmidrule(l){4-7}
\ck   & \xk   & \xk   & 94.5  & 94.8  & 95.4  & 95.8 \\
\cmidrule(r){1-3} \cmidrule(l){4-7}
\ck   & \ck   & \xk   & \underline{95.1}  & \underline{95.7}  & \textbf{96.3}  & \textbf{96.6} \\
\ck   & \ck   & \ck   & \textbf{95.2}  & \textbf{96.0}  & \underline{96.2}  & \underline{96.5} \\
\bottomrule
\end{tabular}%

        \end{adjustbox}
        \vspace{-0.05in}
        \caption{$k$-shot AS ablations: MVTec-AD. Bold/underline indicate the best/runner-up.}\label{tab:winclipp_al}
        \vspace{0.1in}
        \begin{adjustbox}{width=0.98\linewidth}
        \begin{tabular}{l|ccc|c}
\toprule
Method & PCB2 & PCB4  & Pipe fryum & Mean \\
\midrule
WinCLIP & 51.2 & 79.6 & 69.7 & 78.1  \\
 + specific states & \textbf{56.5}  & \textbf{82.7}   & \textbf{70.4} & \textbf{78.9} \\
\bottomrule
\end{tabular}%
        \end{adjustbox}
        \vspace{-0.05in}
        \caption{Ablation on specific states: VisA.}\label{ablate:context_words}
    \end{minipage}
    \hfill
    \vspace{-0.18in}
\end{table*}
\begin{table}[t]
    \centering
    \begin{adjustbox}{width=0.83\linewidth}
\begin{tabular}{l|ccc|r}
\toprule
Method & pAUROC & PRO   & $F_1$-max & \multicolumn{1}{c}{Time (ms)} \\
\midrule
Patch-token & 22.4  & 2.3   & 8.0   & \textbf{95.5\dev{18.8}} \\
Image tiling & 77.9  & 57.5  & 25.5  & 1442.1\dev{62.2} \\
\midrule
\textbf{WinCLIP (ours)} & \textbf{85.1} & \textbf{64.6} & \textbf{31.7} & 389.4\dev{18.5} \\
  w/o image-scale & {82.0}  & {63.0}  & {29.5} & 378.6\dev{20.2} \\
 w/o mid-scale & {84.0}  & {61.6}  & {30.5} & \underline{190.7}\dev{13.9} \\
 w/o small-scale & \underline{84.7}  & \underline{63.6}  & \underline{30.6} & 265.4\dev{15.9} \\
 w/o Harmonic avg. & 81.5  & 60.5  & 27.3  & 279.9\dev{22.8} \\
\bottomrule
\end{tabular}%

    \end{adjustbox}
    \vspace{-0.05in}
    \caption{Comparison of AS performance on MVTec-AD and its per-image inference time, measured at Amazon EC2 G4dn instances. }\label{tab:al_0shot}
    \vspace{-0.2in}
\end{table}

\subsection{Ablation study}
\label{sec:ablation}

We perform component-wise analysis on MVTec-AD \cite{bergmann2019mvtec}. 
A further study, \eg comparison with CLIP-based PatchCore, effect of different backbones, discussion on failure cases, \textit{etc.}, can be found in the supplementary material.

\vspace{0.05in}
\noindent\textbf{WinCLIP for AC: } 
In Table~\ref{tab:ad_0shot}, we report the individual effect of components that constitute our zero-shot AC model. Firstly, we observe (a) the textual supervision for the word ``\verb|anomalous|'' is crucial to achieve a reasonable performance (``One-class''; Section~\ref{sec:zeroshot}), suggesting the effectiveness of CLIP knowledge about ``abnormality''. Next, we confirm that having a diversity in both (b) state-level and (c) prompt-level texts are the key source of gains. And we remark the proposed state ensemble as a more significant component. 
Finally, we observe (d) applying multi-crop prediction \cite{hinton2012imagenet} could also yield a minor improvement.

\vspace{0.05in}
\noindent\textbf{WinCLIP for AS: }
Table~\ref{tab:al_0shot} validates not only the efficiency of WinCLIP to extract local features for zero-shot AS, but also the effectiveness of multi-scale and harmonic averaging to boost the results. To this end, we consider the following additional baselines that also extract patch-level features: (\textit{i}) \emph{Patch-token} (Section \ref{sec:winclip}): it takes the patch features at the last layer, and (\textit{ii}) \emph{Image tiling}: it first performs dense ``tiling'' on an image and then obtains ``tile'' embeddings for segmentation by forwarding each tile with resizing. Overall, the comparison shows that patch-tokens are not aligned with language despite its fast inference time, while ``Image tiling'' makes a significant computational overhead although it does benefit from their local features. WinCLIP achieves accelerated inference due to its window-based computation of local features, with even better performance. Also based on the multi-scale study, we observe that segmentation benefits from both features with image-level, and middle/local context. Note that the scores from last patch embeddings of ViT encodes global context thanks to self-attention, which contributes to a comprehensive localization in WinCLIP.

\vspace{0.05in}
\noindent\textbf{WinCLIP+ for AC and AS: } 
We ablate on different factors to define WinCLIP+ scores for AC \eqref{eq:winplus_ad} and AS \eqref{eq:winplus_al} respectively. For AC, from Table~\ref{tab:winclipp_ad}, we clearly remark the effectiveness of $\mathrm{ascore}_0$ upon $\max \mathbf{M}^{\tt W}$. Interestingly, we observe $\mathrm{ascore}_0$ is beneficial even in higher-shot regimes where $\max \mathbf{M}^{\tt W}$ can be better, confirming their complementary effects. For AS, in Table~\ref{tab:winclipp_al}, we notice the effect of adding $\mathbf{M}_{\tt m}^{\tt W}$ (or $\mathbf{M}_{\tt s}^{\tt W}$) upon $\mathbf{M}^{\tt P}$, \ie the prediction from WinCLIP features: apart from the good performance of $\mathbf{M}^{\tt P}$, $\mathbf{M}^{\tt W}$ could still provide useful information from its local-awareness.

\vspace{0.05in}
\noindent\textbf{WinCLIP with task-specific defects: } 
As mentioned in Section~\ref{sec:zeroshot}, besides using the generic state words and templates (Fig.~6 of supplementary) to cover common cases, our compositional prompt ensemble also supports task-specific state words, \eg ``missing part'' on PCB/``burnt'' pipe fryum; both VisA and MVTec-AD release specific defect types. Ablation study in Table \ref{ablate:context_words} shows that specific state words further improve zero-shot classification in VisA by $0.8\%$ average AUROC with $5.3\%$ gain on the challenging PCB2. 





\section{Conclusion}

We propose a novel framework to define normality and anomaly via both fine-grained textual definitions and normal reference images for comprehensive anomaly classification and segmentation. First, we show that the CLIP pre-trained on large-scale web data provides a powerful representation with good alignment between texts and images for anomaly recognition tasks. The compositional prompt ensemble defines the normality and anomaly in text and helps to distill knowledge from the pre-trained CLIP for better zero-shot anomaly recognition. WinCLIP efficiently aggregates multi-scale features with image-text alignment from window and image-level to perform zero-shot segmentation. Moreover, given a few normal samples, vision based reference association provides complementary information about the two states to language definitions, leading to few-shot WinCLIP+. In recent benchmarks, WinCLIP and WinCLIP+ outperform state-of-the-arts in zero-/few-shot setups with considerable margins. We believe our work will bring values complementary to standard one-class methods. For further improvement, vision-language pre-training with industrial domain data is a promising direction that is left as a future work. 

{\small
\bibliographystyle{ieee_fullname}
\bibliography{egbib}
}

\clearpage
\appendix
\onecolumn
\FloatBarrier

\begin{center}
    {\bf {\LARGE Supplementary Material}}
\end{center}
\begin{center}
{\bf {\Large WinCLIP: Zero-/Few-Shot Anomaly Classification and Segmentation}}
\end{center}

\setcounter{figure}{5}
\setcounter{table}{7}

\section{Experimental details}


\noindent\textbf{Compositional prompt ensemble. }
Figure~\ref{fig:comp_prompt} provides a detailed list of prompts we adopt to perform compositional prompt ensemble proposed in Section~4.1 of the main text. 
Recall that we consider two levels of prompts: \ie (a) state-level, and (b) template level. A complete prompt can be composed by replacing the token \verb|[c]| in a template-level prompt with one of state-level prompt, either from the normal or anomaly states. Each of the state-level prompt takes an object-level label \verb|[o]|. In our experiments, we use the object name words available for both MVTec-AD and VisA per dataset to replace \verb|[o]|.

\begin{figure*}
\noindent\begin{minipage}[t]{0.32\linewidth}
(a) \emph{State}-level (normal)

{\tt \small
\begin{itemize}
    \item c := "[o]"
    \item c := "flawless [o]"
    \item c := "perfect [o]"
    \item c := "unblemished [o]"
    \item c := "[o] without flaw"
    \item c := "[o] without defect"
    \item c := "[o] without damage"
\end{itemize}
}

(b) \emph{State}-level (anomaly)

{\tt \small
\begin{itemize}
    \item c := "damaged [o]"
    \item c := "[o] with flaw"
    \item c := "[o] with defect"
    \item c := "[o] with damage"
\end{itemize}
}
\end{minipage}
\hfill
\begin{minipage}[t]{0.33\linewidth}
(c) \emph{Template}-level

{\tt \small
\begin{itemize}
\item "a cropped photo of the [c]."
\item "a cropped photo of a [c]."
\item "a close-up photo of a [c]."
\item "a close-up photo of the [c]."
\item "a bright photo of a [c]." 
\item "a bright photo of the [c]."
\item "a dark photo of the [c]."
\item "a dark photo of a [c]."
\item "a jpeg corrupted photo of a [c]."
\item "a jpeg corrupted photo of the [c]."
\end{itemize}
}
\end{minipage}
\begin{minipage}[t]{0.33\linewidth}
{\tt \small
\begin{itemize}
\item {\rm (cont'd)} "a blurry photo of the [c]."
\item "a blurry photo of a [c]."
\item "a photo of a [c]."
\item "a photo of the [c]."
\item "a photo of a small [c]."
\item "a photo of the small [c]."
\item "a photo of a large [c]."
\item "a photo of the large [c]."
\item "a photo of the [c] for visual inspection."
\item "a photo of a [c] for visual inspection."
\item "a photo of the [c] for anomaly detection."
\item "a photo of a [c] for anomaly detection."
\end{itemize}
}
\end{minipage}
\caption{Lists of multi-level prompts considered in this paper to construct compositional prompt ensemble.}
\label{fig:comp_prompt}
\end{figure*}

\vspace{0.05in}
\noindent\textbf{Data pre-processing. }
For CLIP-based models, including our proposed WinCLIP and WinCLIP+, we apply the data pre-processing pipeline given in OpenCLIP \cite{ilharco_gabriel_2021_5143773} for both MVTec-AD and VisA datasets to minimize potential train-test discrepancy. Specifically, it performs a channel-wise standardization with the pre-computed mean \texttt{[0.48145466, 0.4578275, 0.40821073]} and standard deviation \texttt{[0.26862954, 0.26130258, 0.27577711]} after normalizing each RGB image into $[0, 1]$, followed by a bicubic re-sizing based on the \verb|Pillow| implementation. By default, we make the input resolution to be $240$ for the shorter edge from the re-sizing, to be compatible with ViT-B/16+ in our experiments. This re-sizing policy also applies to other baseline models for fairer comparisons, although we keep the remaining parts of their original data pre-processing pipelines. In addition, similar policy can also be used in other backbones with input of different resolutions.

\vspace{0.05in}
\noindent\textbf{Evaluation metrics. }
Although the AUROC is a good metric for balanced dataset, it provides an inflated view of model performance in imbalanced dataset, especially in anomaly segmentation where the normal pixels dominate anomalies. This is also discussed by Zou et al.~\cite{zou2022spot}. $F_1$-max is computed from the precision and recall for the anomalous samples at the optimal threshold, which is a more straightforward metric to measure the upper bound of anomaly prediction performance across thresholds. Thus we acknowledge that the low-shot anomaly segmentation is still not solved since our best model only achieves $<60\%$ $F_1$-max for both MVTec-AD and VisA, even though WinCLIP+ achieves $>95\%$ pixel-AUROC. In addition, our WinCLIP and WinCLIP+ outperform all the compared methods in terms of all these metrics on the setups, demonstrating the effectiveness of the proposed methods.

\vspace{0.05in}
\noindent\textbf{Other implementation details.}
(\textit{i}) The ViT-B/16+ architecture \cite{ilharco_gabriel_2021_5143773}, that we mainly adopt in our experiments, is a modification of ViT-B/16 \cite{touvron2021training} with (a) an increased dimension in both image ($768 \rightarrow 896$) and text ($512 \rightarrow 640$) embeddings, as well as in (b) the input resolution ($224^2 \rightarrow 240^2$; $196 \rightarrow 225$ tokens);
(\textit{ii}) We note that CLIP models require the square-shaped resolution, \eg $240^2$ for ViT-B/16+, to be compatible with the attention layers inside. 
Although the MVTec-AD benchmark already consists of square images, most of images in the VisA benchmark are non-squared (\eg $1500\times1000$) and simply taking a crop can affect the anomaly status of the given images. 
In this respect, to enable CLIP-based models properly handle non-squared images in our experiments, we perform a simple ``image tiling'' scheme. Specifically, for such non-squared images, we first extract multiple overlapping (squared) ``tiles'' of size the shorter edge $L_s$, by taking a sliding window across the longer edge. Then we average the predictions from the tiles to get the final (either in image- and pixel-level) prediction. The stride for the sliding is set to $0.8\cdot L_s$ at most, \ie the tiles have overlaps with its neighbors at least in $0.2\cdot L_s$; (\textit{iii}) In addition, for the baseline results, we use our re-implementation of SPADE \cite{cohen2020sub} and PaDiM \cite{defard2021padim}, and adopt the official implementation of PatchCore\footnote{\url{https://github.com/amazon-science/patchcore-inspection}} in our experiments. 


\section{Additional results on ablation study}


\begin{table*}[t]
    \centering
    \hfill
    \begin{minipage}[t]{0.71\linewidth}
        \centering
        \begin{adjustbox}{width=\linewidth}
\begin{tabular}{clccccccc}
\toprule
\multicolumn{2}{l}{MVTec-AD (few-shot)} &       & \multicolumn{3}{c}{Anomaly classification} & \multicolumn{3}{c}{Anomaly segmentation} \\
\cmidrule(r){4-6} \cmidrule(l){7-9}
Setup & Method & Backbone & pAUROC & PRO   & $F_1$-max & AUROC & AUPR  & $F_1$-max \\
\midrule
\multirow{4}[6]{*}{1-shot} & PatchCore \cite{roth2022towards} & WRN-50-2 & 83.4\dev{3.0} & 92.2\dev{1.5} & 90.5\dev{1.5} & 92.0\dev{1.0} & 79.7\dev{2.0} & 50.4\dev{2.1} \\
\cmidrule{2-3}      & PatchCore (hidden) & \multirow{3}[4]{*}{ViT-B/16+} & 79.9\dev{4.8} & 88.6\dev{2.7} & 88.7\dev{1.1} & 91.8\dev{1.0} & 74.1\dev{2.3} & 47.6\dev{2.6} \\
      & PatchCore (last) &       & 83.3\dev{3.8} & 90.7\dev{2.1} & 89.8\dev{1.4} & 92.3\dev{0.9} & 74.5\dev{2.2} & 47.7\dev{2.9} \\
\cmidrule{2-2}      & \textbf{WinCLIP+ (ours)} &       & \textbf{92.7\dev{1.9}} & \textbf{96.7\dev{0.7}} & \textbf{93.5\dev{1.0}} & \textbf{95.2\dev{0.5}} & \textbf{87.1\dev{1.2}} & \textbf{55.9\dev{2.7}} \\
\midrule
\multirow{4}[6]{*}{2-shot} & PatchCore \cite{roth2022towards} & WRN-50-2 & 86.3\dev{3.3} & 93.8\dev{1.7} & 92.0\dev{1.5} & 93.3\dev{0.6} & 82.3\dev{1.3} & 53.0\dev{1.7} \\
\cmidrule{2-3}      & PatchCore (hidden) & \multirow{3}[4]{*}{ViT-B/16+} & 84.1\dev{2.9} & 90.7\dev{1.9} & 90.2\dev{1.2} & 93.5\dev{0.7} & 77.9\dev{1.8} & 51.4\dev{2.1} \\
      & PatchCore (last) &       & 86.5\dev{2.5} & 92.3\dev{1.4} & 91.1\dev{1.6} & 93.1\dev{0.9} & 76.8\dev{2.0} & 49.8\dev{2.2} \\
\cmidrule{2-2}      & \textbf{WinCLIP+ (ours)} &       & \textbf{94.0\dev{1.7}} & \textbf{97.0\dev{0.7}} & \textbf{94.0\dev{1.0}} & \textbf{96.0\dev{0.3}} & \textbf{88.4\dev{0.9}} & \textbf{58.4\dev{1.7}} \\
\midrule
\multirow{4}[6]{*}{4-shot} & PatchCore \cite{roth2022towards} & WRN-50-2 & 88.8\dev{2.6} & 94.5\dev{1.5} & 92.6\dev{1.6} & 94.3\dev{0.5} & 84.3\dev{1.6} & 55.0\dev{1.9} \\
\cmidrule{2-3}      & PatchCore (hidden) & \multirow{3}[4]{*}{ViT-B/16+} & 87.5\dev{3.1} & 92.5\dev{1.8} & 91.7\dev{1.4} & 94.7\dev{0.6} & 81.2\dev{1.6} & 54.4\dev{1.9} \\
      & PatchCore (last) &       & 89.9\dev{2.2} & 94.2\dev{1.4} & 92.7\dev{1.1} & 94.0\dev{0.6} & 78.9\dev{1.6} & 52.2\dev{1.5} \\
\cmidrule{2-2}      & \textbf{WinCLIP+ (ours)} &       & \textbf{94.8\dev{1.5}} & \textbf{97.5\dev{0.7}} & \textbf{94.2\dev{0.9}} & \textbf{96.2\dev{0.3}} & \textbf{89.0\dev{0.8}} & \textbf{59.5\dev{1.8}} \\
\bottomrule
\end{tabular}%

        \end{adjustbox}
        \caption{Comparison of few-shot performances on MVTec-AD. We report the mean and standard deviation over 5 random seeds for each measurement. Bold indicates the best performance.}
        \label{tab:ab_patchcore}
    \end{minipage}
    \hfill
    \begin{minipage}[t]{0.24\linewidth}
        \centering
        \small
        \begin{adjustbox}{width=\linewidth}
\begin{tabular}{cc|cc}
\toprule
\multicolumn{4}{c}{MVTec-AD (zero-shot)} \\
\midrule
Model & Size  & AC    & AS \\
\midrule
RN50  & $224^2$ & 79.8  & 65.6 \\
RN101 & $224^2$ & 79.2  & 62.4 \\
RN50x4 & $288^2$ & 81.9  & 71.3 \\
RN50x16 & $384^2$ & 82.3  & 65.3 \\
ViT-B/16 & $224^2$ & 86.1  & 71.1 \\
ViT-B/16+ & $240^2$ & 90.8  & 85.1 \\
ViT-L/14 & $224^2$ & 86.1  & 64.4 \\
\bottomrule
\end{tabular}%

        \end{adjustbox}
        \caption{Comparison of WinCLIP performance in AUROC (for AC) and pAUROC (for AS) on zero-shot MVTec-AD, across different CLIP backbone architectures. }
        \label{tab:ab_arch}
    \end{minipage}
    \hfill
    \vspace{-0.05in}
\end{table*}

\noindent\textbf{Comparison with CLIP-based PatchCore: } 
PatchCore \cite{roth2022towards}, a current state-of-the-art considered in our experiments, is originally based on the internal features of convolutional network: \eg WideResNet-50-2 (WRN-50-2)  \cite{zagoruyko2016wideresnet} pre-trained on ImageNet.   
In Table~\ref{tab:ab_patchcore}, we test whether PatchCore can further benefit from the CLIP-based backbone that our WinCLIP+ is based on. Specifically, we additionally consider two variants of PatchCore that take the patch-token features of CLIP-based ViT-B/16+ backbone, one from (a) the 6$^\text{th}$- and 9$^\text{th}$-layer of ViT (which corresponds to \texttt{block2} and \texttt{block3} in ResNet-like models as considered by \cite{roth2022towards}; ``\emph{hidden}''), and the other one from (b) the last layer of ViT (``\emph{last}''). Overall, we have the following observations. First, in case of the ViT-B/16+ backbone, PatchCore performs better with the last layer, which is in contrast to the cases of convolutional backbones. Second, compared to the original PatchCore, the CLIP-based variants achieve no better performances. Third, WinCLIP+ significantly outperforms ``PatchCore (last)'' where our WinCLIP+ also utilizes the last patch-token features, namely as referred as $\mathbf{F}^{\tt P}$ (Section~4.3 of the main text). 
The results confirm the effectiveness of (a) our simple association-based module over a more sophisticated PatchCore\footnote{Technically, PatchCore incorporates several techniques upon a patch-level memory scheme, \eg local patch aggregation, clustering and score re-weighting.} in ViT, and (b) the WinCLIP features $\mathbf{F}^{\tt W}$.

\vspace{0.05in}
\noindent\textbf{Effect of different CLIP backbones: }
Table~\ref{tab:ab_arch}, on the other hand, explores the effect of different CLIP architectures to the WinCLIP zero-shot performance. Specifically, on zero-shot setups, we compare AUROC (and pixel-AUROC) from WinCLIP in AC (and AS) testing over the CLIP pre-trained models available at OpenCLIP,\footnote{\url{https://github.com/mlfoundations/open_clip}} including our default choice of ViT-B/16+. To apply WinCLIP for ResNet-based backbones, we notice that the CLIP implementation of ResNet architectures incorporates an attention layer to perform the feature pooling, namely as \emph{attention pooling}, similar to ViT-based architectures. In this respect, for the CLIP-ResNet models, we apply our window-based inference to perform zero-shot AS from the convolutional feature map before the attention pooling, in the same way of applying WinCLIP for ViTs. Here, we remark that the effective patch size of each pixel on the last feature map (before the pooling) of ResNet-based models is designed to be $32$ (the downsampling rate), which is larger than those of ViTs we test, \eg of 16. Overall, we observe that ViT-based models generally show better performance compared to ResNets, in both AC and AS. The particular gap in AS is possibly due to the bigger patch sizes in ResNets, which can result in more blurry outputs. Still, we observe the performance benefits from larger models or resolutions in both types of architecture.

\section{Additional qualitative results}

\begin{figure*}[ht]
  \includegraphics[width=\linewidth]{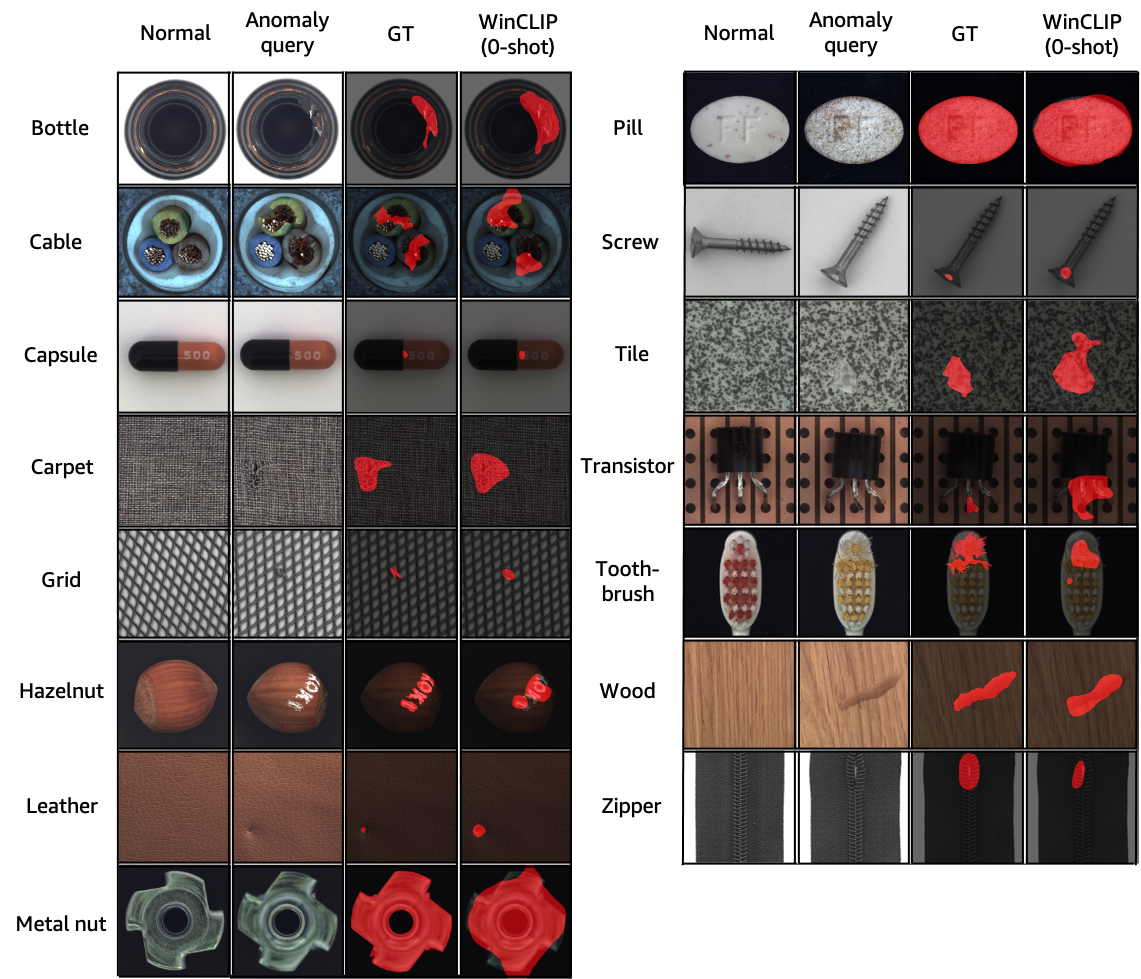}
  \caption{Additional qualitative results from WinCLIP (0-shot), tested on MVTec-AD.}
  \label{fig:good_0shot_mvtec}
\end{figure*}

\begin{figure*}[ht]
  \includegraphics[width=\linewidth]{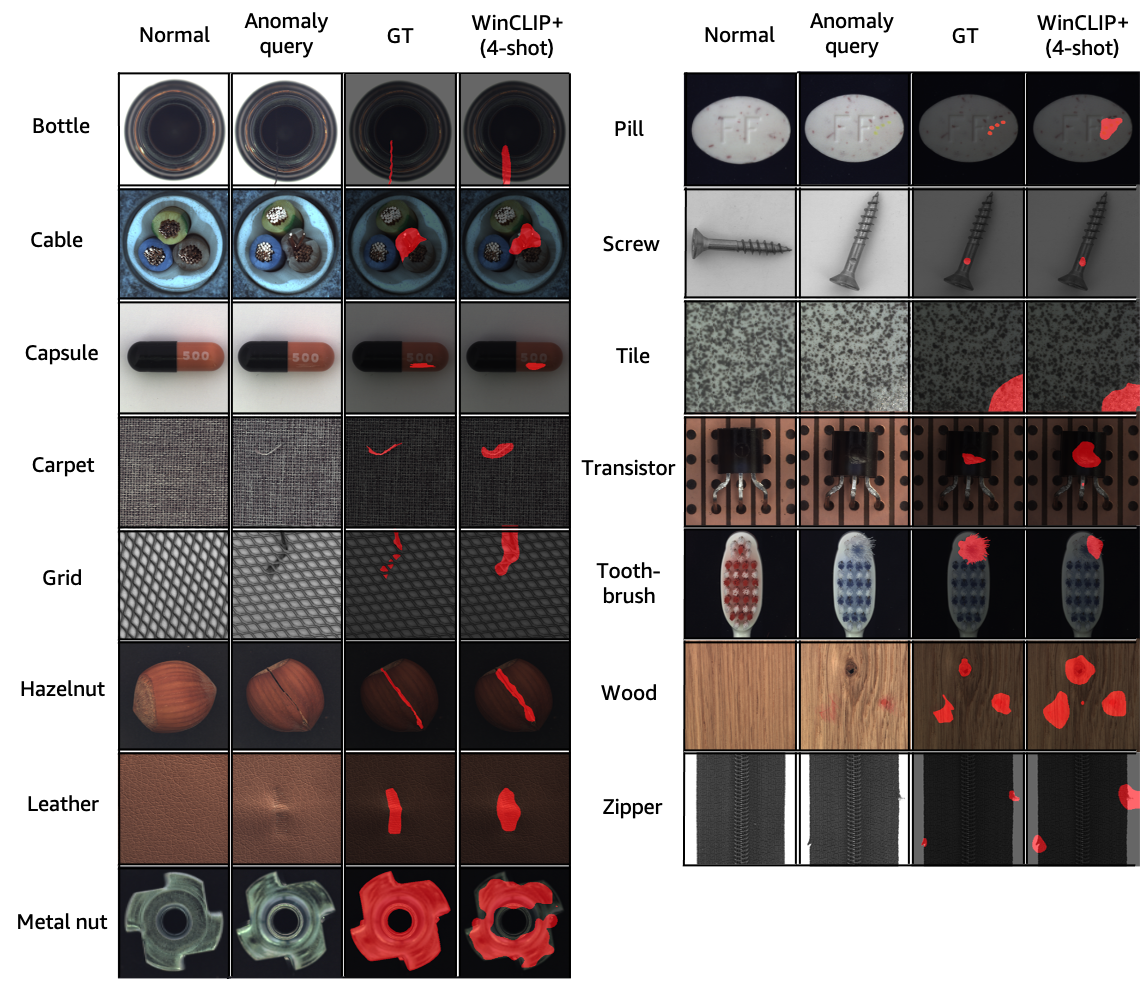}
  \caption{Additional qualitative results from few-shot WinCLIP+ (4-shot), tested on MVTec-AD.}
  \label{fig:good_4shot_mvtec}
\end{figure*}

\begin{figure*}[ht]
  \includegraphics[width=\linewidth]{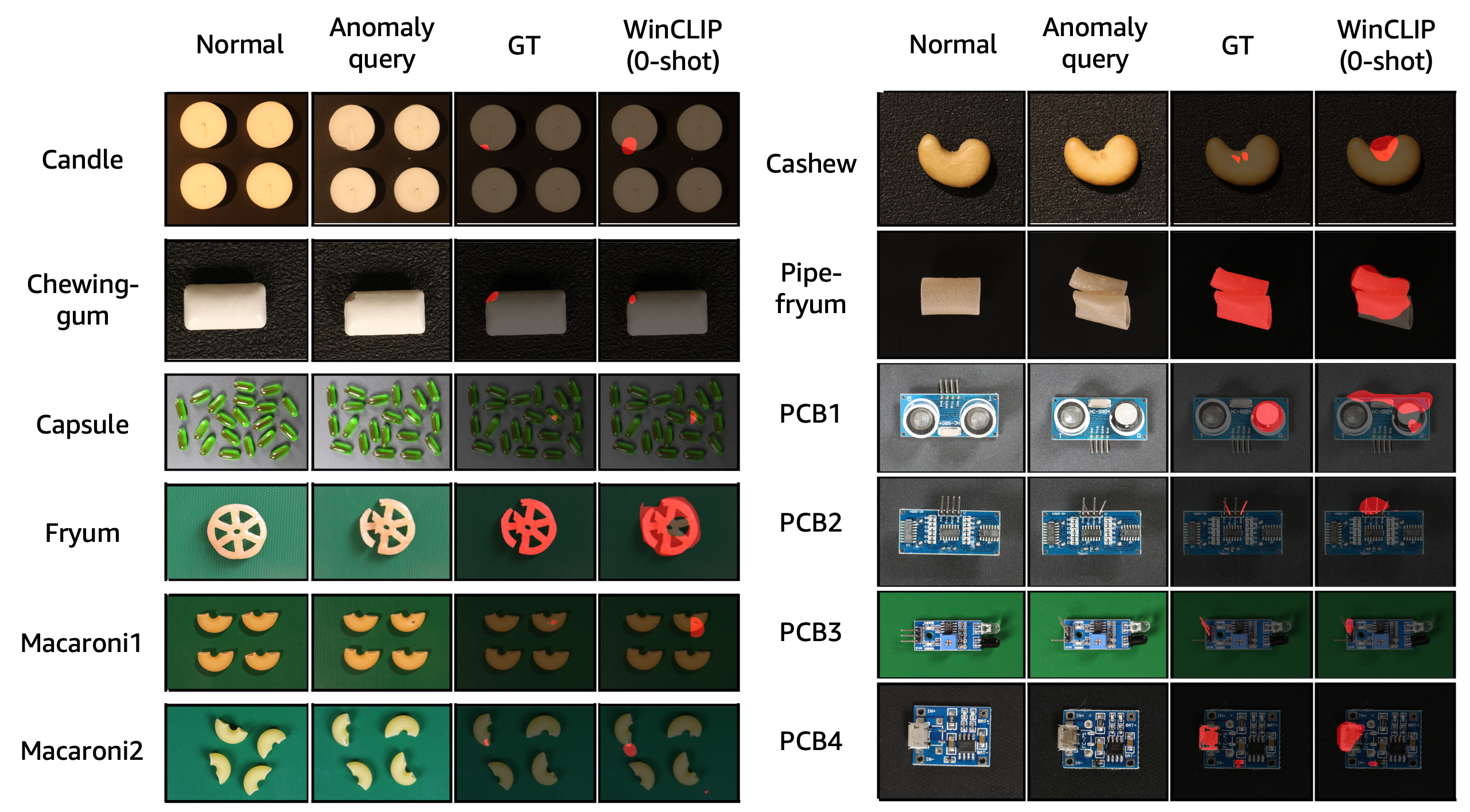}
  \caption{Additional qualitative results from WinCLIP (0-shot), tested on VisA.}
  \label{fig:good_0shot_visa}
  \vspace{0.1in}
  \includegraphics[width=\linewidth]{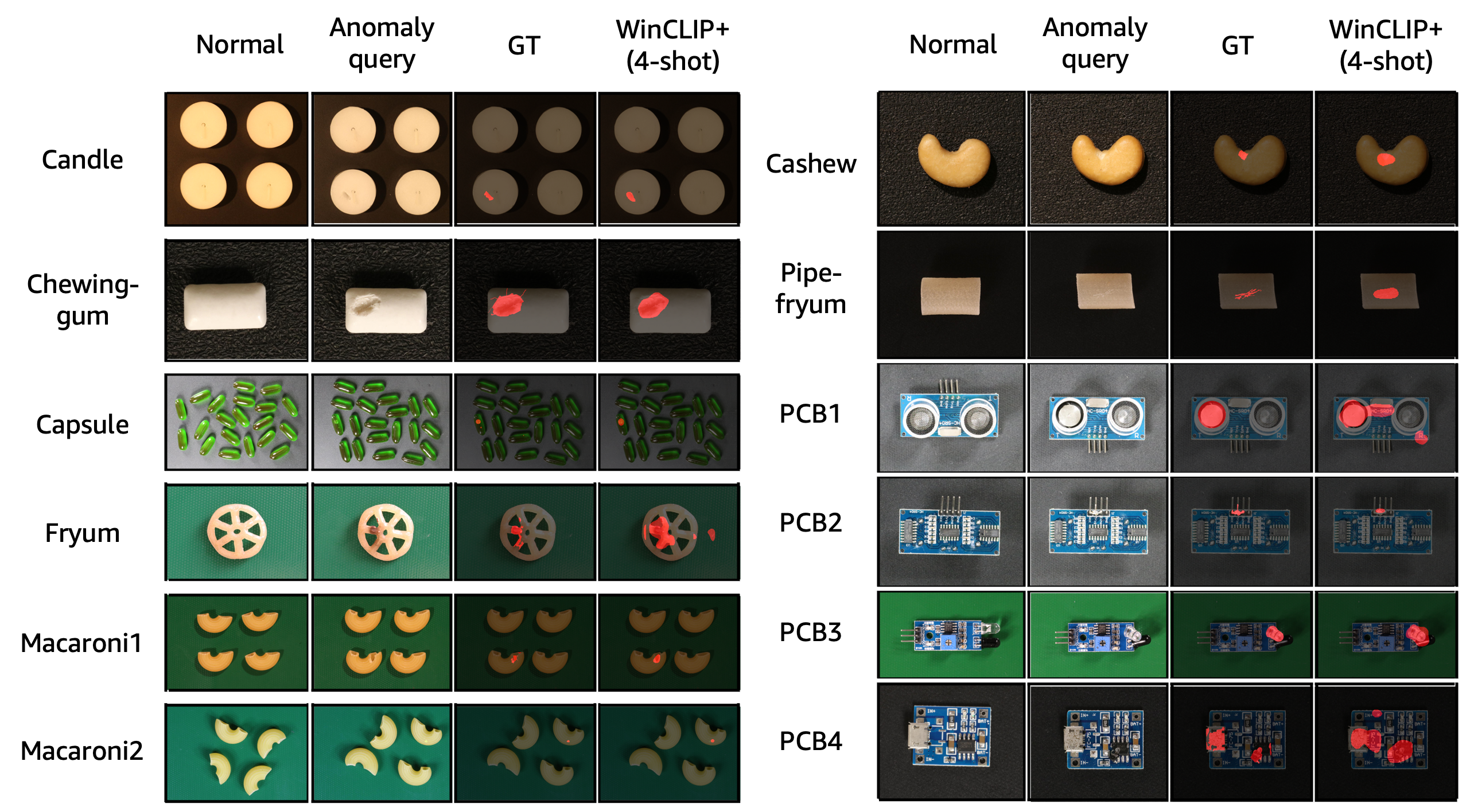}
  \caption{Additional qualitative results from few-shot WinCLIP+ (4-shot), tested on VisA.}
  \label{fig:good_4shot_visa}
\end{figure*}

In Figure~\ref{fig:good_0shot_mvtec}-\ref{fig:good_4shot_visa}, we provide further qualitative results obtained from our (zero-shot) WinCLIP and (few-shot) WinCLIP+ for anomaly segmentation, both in MVTec-AD and VisA considered in our experiments. Specifically, we report MVTec-AD results in Figure~\ref{fig:good_0shot_mvtec} and \ref{fig:good_4shot_mvtec}, and VisA results in Figure~\ref{fig:good_0shot_visa} and \ref{fig:good_4shot_visa}.

\begin{figure*}[ht]
  \includegraphics[width=\linewidth]{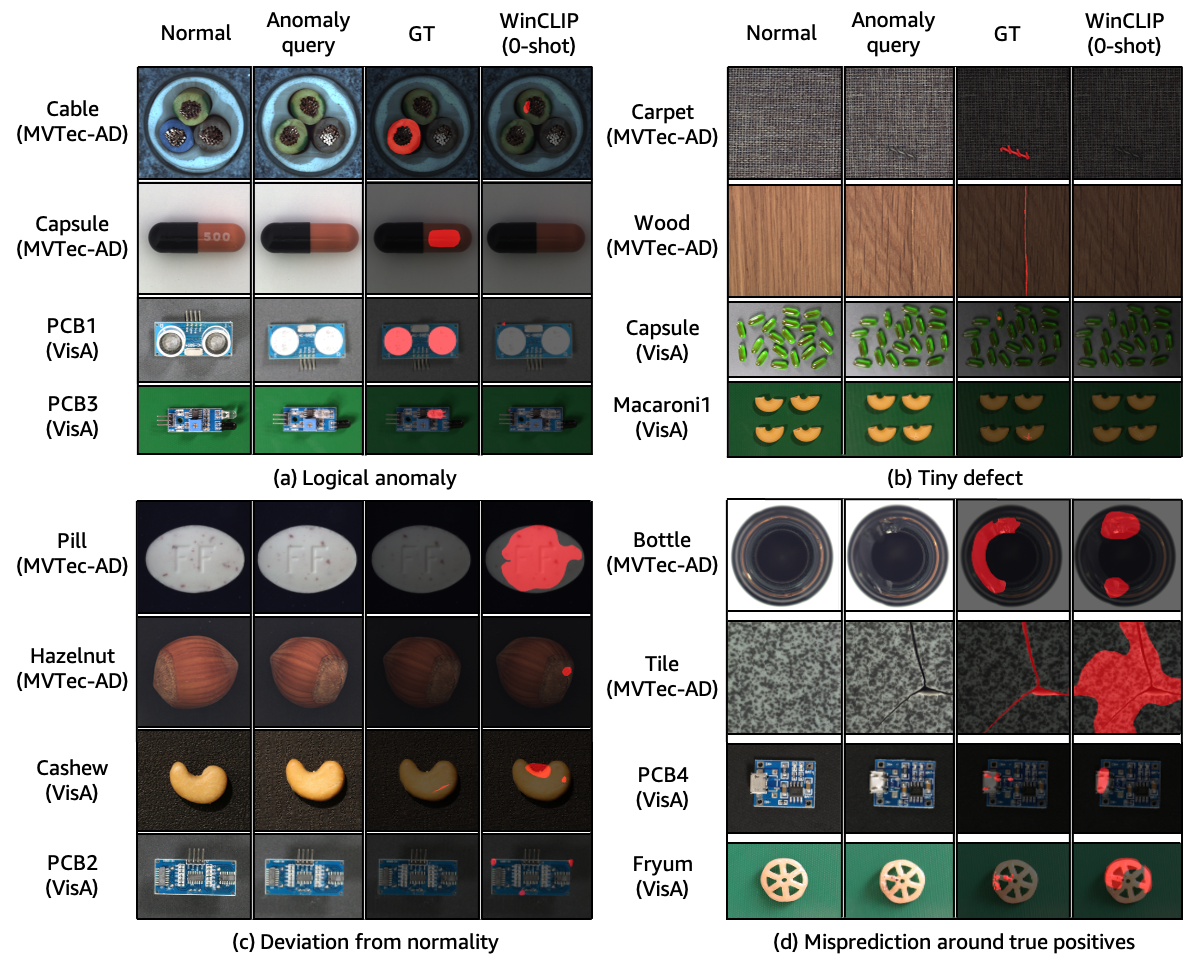}
  \caption{Curated illustrations of failure cases from zero-shot WinCLIP.}
  \label{fig:fail_0shot}
\end{figure*}

\clearpage
\noindent\textbf{Failure cases. } 
We present some failure examples from both MVTec-AD and VisA for language driven zero-shot WinCLIP in Figure \ref{fig:fail_0shot}. Note that the normal images are shown just for better illustration and are not used in model prediction. The first major factor causing the failure is the logical anomaly \cite{bergmann2022beyond} illustrated in Figure \ref{fig:fail_0shot}\textcolor{red}{(a)}, \eg misplaced axis in cable, missing text on capsule, missing capacitor in PCB1 and bent component in PCB3. Such type of anomalies need to be clarified by normal reference images while language might be not sufficient. The issues are alleviated by our few-normal-shot WinCLIP+. The second major factor refers to tiny defect illustrated in Figure \ref{fig:fail_0shot}\textcolor{red}{(b)}, such as the ones in carpet, wood, capsule, macaroni1. We conjecture that spatial features with more local details might improve these cases, which is left for future exploration. The third major factor is the irrelevant deviation from normality that are not defects of interests illustrated in Figure \ref{fig:fail_0shot}\textcolor{red}{(c)}, \eg the tiny red/white dots in pill/hazelnut, extra ingredient on cashew, designed holes and acceptable scratches in PCB2. We hypothesize that more clarification on these deviation and a pre-trained model with better understanding on these states might alleviate the problem. Lastly, although WinCLIP can roughly localize anomalies such as the cases in bottle, tile, PCB4 and fryum, it makes some errors around the true positives, illustrated in Figure \ref{fig:fail_0shot}\textcolor{red}{(d)}. However, we argue this is minor as the rough anomaly localization is sufficient to explain where the defects are for visual inspection. 

\clearpage
\FloatBarrier
\section{Detailed quantitative results}

In this section, we report the detailed, subset-level performance values for the evaluation metrics provided in Table~1 and 4 of the main text. 
Specifically, we report MVTec-AD results in Table~\ref{tab:mvtec/ac/roc}-\ref{tab:mvtec/as/pf1} and VisA results in Table~\ref{tab:visa/ac/roc}-\ref{tab:visa/as/pf1}.


\begin{table*}[ht]
  \centering
  \begin{adjustbox}{width=\linewidth}
\begin{tabular}{cccccccccccccc}
\toprule
MVTec-AD (AC) & $K=0$ & \multicolumn{4}{c}{$K=1$}     & \multicolumn{4}{c}{$K=2$}     & \multicolumn{4}{c}{$K=4$} \\
\cmidrule(lr){2-2} \cmidrule(lr){3-6} \cmidrule(lr){7-10} \cmidrule(l){11-14}
AUROC & \textbf{WinCLIP} & SPADE & PaDiM & PatchCore & \textbf{WinCLIP+} & SPADE & PaDiM & PatchCore & \textbf{WinCLIP+} & SPADE & PaDiM & PatchCore & \textbf{WinCLIP+} \\
\cmidrule(r){1-1} \cmidrule(lr){2-2} \cmidrule(lr){3-6} \cmidrule(lr){7-10} \cmidrule(l){11-14}
Bottle & 99.2\dev{0.0} & 98.7\dev{0.6} & 97.4\dev{0.7} & 99.4\dev{0.4} & 98.2\dev{0.9} & 99.5\dev{0.1} & 98.5\dev{1.0} & 99.2\dev{0.3} & 99.3\dev{0.3} & 99.5\dev{0.2} & 98.8\dev{0.2} & 99.2\dev{0.3} & 99.3\dev{0.4} \\
Cable & 86.5\dev{0.0} & 71.2\dev{3.3} & 57.7\dev{4.6} & 88.8\dev{4.2} & 88.9\dev{1.9} & 76.2\dev{5.2} & 62.3\dev{5.9} & 91.0\dev{2.7} & 88.4\dev{0.7} & 83.4\dev{3.1} & 70.0\dev{6.1} & 91.0\dev{2.7} & 90.9\dev{0.9} \\
Capsule & 72.9\dev{0.0} & 70.2\dev{3.0} & 57.7\dev{7.3} & 67.8\dev{2.9} & 72.3\dev{6.8} & 70.9\dev{6.1} & 64.3\dev{3.0} & 72.8\dev{7.0} & 77.3\dev{8.8} & 78.9\dev{5.5} & 65.2\dev{2.5} & 72.8\dev{7.0} & 82.3\dev{8.9} \\
Carpet & 100.0\dev{0.0} & 98.1\dev{0.2} & 96.6\dev{1.0} & 95.3\dev{0.8} & 99.8\dev{0.3} & 98.3\dev{0.4} & 97.8\dev{0.5} & 96.6\dev{0.5} & 99.8\dev{0.3} & 98.6\dev{0.2} & 97.9\dev{0.4} & 96.6\dev{0.5} & 100.0\dev{0.0} \\
Grid  & 98.8\dev{0.0} & 40.0\dev{6.8} & 54.2\dev{6.7} & 63.6\dev{10.3} & 99.5\dev{0.3} & 41.3\dev{3.6} & 67.2\dev{4.2} & 67.7\dev{8.3} & 99.4\dev{0.2} & 44.6\dev{6.6} & 68.1\dev{3.8} & 67.7\dev{8.3} & 99.6\dev{0.1} \\
Hazelnut & 93.9\dev{0.0} & 95.8\dev{1.3} & 88.3\dev{2.6} & 88.3\dev{2.7} & 97.5\dev{1.4} & 96.2\dev{2.1} & 90.8\dev{0.8} & 93.2\dev{3.8} & 98.3\dev{0.7} & 98.4\dev{1.3} & 91.9\dev{1.2} & 93.2\dev{3.8} & 98.4\dev{0.4} \\
Leather & 100.0\dev{0.0} & 100.0\dev{0.0} & 97.5\dev{0.7} & 97.3\dev{0.7} & 99.9\dev{0.0} & 100.0\dev{0.0} & 97.5\dev{0.9} & 97.9\dev{0.7} & 99.9\dev{0.0} & 100.0\dev{0.0} & 98.5\dev{0.2} & 97.9\dev{0.7} & 100.0\dev{0.0} \\
Metal nut & 97.1\dev{0.0} & 71.0\dev{2.2} & 53.0\dev{3.8} & 73.4\dev{2.9} & 98.7\dev{0.8} & 77.0\dev{7.9} & 54.8\dev{3.8} & 77.7\dev{8.5} & 99.4\dev{0.2} & 77.8\dev{5.7} & 60.7\dev{5.2} & 77.7\dev{8.5} & 99.5\dev{0.2} \\
Pill  & 79.1\dev{0.0} & 86.5\dev{3.1} & 61.3\dev{3.8} & 81.9\dev{2.8} & 91.2\dev{2.1} & 84.8\dev{0.9} & 59.1\dev{6.4} & 82.9\dev{2.9} & 92.3\dev{0.7} & 86.7\dev{0.3} & 54.9\dev{2.7} & 82.9\dev{2.9} & 92.8\dev{1.0} \\
Screw & 83.3\dev{0.0} & 46.7\dev{2.5} & 55.0\dev{2.5} & 44.4\dev{4.6} & 86.4\dev{0.9} & 46.6\dev{2.2} & 54.0\dev{4.4} & 49.0\dev{3.8} & 86.0\dev{2.1} & 50.5\dev{5.4} & 50.0\dev{4.1} & 49.0\dev{3.8} & 87.9\dev{1.2} \\
Tile  & 100.0\dev{0.0} & 99.9\dev{0.1} & 92.2\dev{2.2} & 99.0\dev{0.9} & 99.9\dev{0.0} & 99.9\dev{0.1} & 93.3\dev{1.1} & 98.5\dev{1.0} & 99.9\dev{0.2} & 100.0\dev{0.0} & 93.1\dev{0.6} & 98.5\dev{1.0} & 99.9\dev{0.1} \\
Toothbrush & 87.5\dev{0.0} & 71.7\dev{2.6} & 82.5\dev{1.2} & 83.3\dev{3.8} & 92.2\dev{4.9} & 78.6\dev{3.2} & 87.6\dev{4.2} & 85.9\dev{3.5} & 97.5\dev{1.6} & 78.8\dev{5.2} & 89.2\dev{2.5} & 85.9\dev{3.5} & 96.7\dev{2.6} \\
Transistor & 88.0\dev{0.0} & 77.2\dev{2.0} & 73.3\dev{6.0} & 78.1\dev{6.9} & 83.4\dev{3.8} & 81.3\dev{3.7} & 72.8\dev{6.3} & 90.0\dev{4.3} & 85.3\dev{1.7} & 81.4\dev{2.1} & 82.4\dev{6.5} & 90.0\dev{4.3} & 85.7\dev{2.5} \\
Wood  & 99.4\dev{0.0} & 98.8\dev{0.3} & 96.1\dev{1.2} & 97.8\dev{0.3} & 99.9\dev{0.1} & 99.2\dev{0.4} & 96.9\dev{0.5} & 98.3\dev{0.6} & 99.9\dev{0.1} & 98.9\dev{0.6} & 97.0\dev{0.2} & 98.3\dev{0.6} & 99.8\dev{0.3} \\
Zipper & 91.5\dev{0.0} & 89.3\dev{1.9} & 85.8\dev{2.7} & 92.3\dev{0.5} & 88.8\dev{5.9} & 93.3\dev{2.9} & 86.3\dev{2.6} & 94.0\dev{2.1} & 94.0\dev{1.4} & 95.1\dev{1.3} & 88.3\dev{2.0} & 94.0\dev{2.1} & 94.5\dev{0.5} \\
\cmidrule(r){1-1} \cmidrule(lr){2-2} \cmidrule(lr){3-6} \cmidrule(lr){7-10} \cmidrule(l){11-14}
Mean  & \textbf{91.8\dev{0.0}} & 81.0\dev{2.0} & 76.6\dev{3.1} & 83.4\dev{3.0} & \textbf{93.1\dev{2.0}} & 82.9\dev{2.6} & 78.9\dev{3.1} & 86.3\dev{3.3} & \textbf{94.4\dev{1.3}} & 84.8\dev{2.5} & 80.4\dev{2.5} & 88.8\dev{2.6} & \textbf{95.2\dev{1.3}} \\
\bottomrule
\end{tabular}%

  \end{adjustbox}
  \caption{Comparison of anomaly classification (AC) performance in terms of class-wise AUROC on MVTec-AD. We report the mean and standard deviation over 5 random seeds for each measurement.}
  \label{tab:mvtec/ac/roc}
  \vspace{0.1in}
  \begin{adjustbox}{width=\linewidth}
\begin{tabular}{cccccccccccccc}
\toprule
MVTec-AD (AC) & $K=0$ & \multicolumn{4}{c}{$K=1$}     & \multicolumn{4}{c}{$K=2$}     & \multicolumn{4}{c}{$K=4$} \\
\cmidrule(lr){2-2} \cmidrule(lr){3-6} \cmidrule(lr){7-10} \cmidrule(l){11-14}
AUPR  & \textbf{WinCLIP} & SPADE & PaDiM & PatchCore & \textbf{WinCLIP+} & SPADE & PaDiM & PatchCore & \textbf{WinCLIP+} & SPADE & PaDiM & PatchCore & \textbf{WinCLIP+} \\
\cmidrule(r){1-1} \cmidrule(lr){2-2} \cmidrule(lr){3-6} \cmidrule(lr){7-10} \cmidrule(l){11-14}
Bottle & 99.8\dev{0.0} & 99.6\dev{0.1} & 99.2\dev{0.2} & 99.8\dev{0.1} & 99.4\dev{0.3} & 99.8\dev{0.0} & 99.6\dev{0.3} & 99.8\dev{0.1} & 99.8\dev{0.1} & 99.9\dev{0.0} & 99.7\dev{0.0} & 99.8\dev{0.1} & 99.8\dev{0.1} \\
Cable & 91.2\dev{0.0} & 79.6\dev{2.3} & 64.9\dev{3.8} & 93.8\dev{2.2} & 93.2\dev{1.1} & 84.5\dev{3.1} & 69.6\dev{6.6} & 95.1\dev{1.3} & 92.9\dev{0.6} & 88.8\dev{1.9} & 76.1\dev{5.6} & 97.1\dev{0.7} & 94.4\dev{0.3} \\
Capsule & 91.5\dev{0.0} & 91.2\dev{0.9} & 86.9\dev{2.2} & 89.4\dev{2.0} & 91.6\dev{2.7} & 91.6\dev{2.1} & 88.4\dev{0.8} & 91.0\dev{2.9} & 93.3\dev{3.6} & 94.4\dev{1.9} & 87.8\dev{0.8} & 94.9\dev{1.1} & 95.1\dev{3.3} \\
Carpet & 100.0\dev{0.0} & 99.4\dev{0.0} & 99.0\dev{0.2} & 98.7\dev{0.2} & 99.9\dev{0.1} & 99.5\dev{0.1} & 99.4\dev{0.1} & 99.0\dev{0.1} & 99.9\dev{0.1} & 99.6\dev{0.1} & 99.4\dev{0.1} & 98.8\dev{0.2} & 100.0\dev{0.0} \\
Grid  & 99.6\dev{0.0} & 66.9\dev{2.1} & 75.0\dev{3.3} & 81.1\dev{4.9} & 99.9\dev{0.1} & 68.3\dev{2.1} & 82.5\dev{2.3} & 84.1\dev{4.0} & 99.8\dev{0.1} & 68.8\dev{4.2} & 83.0\dev{1.8} & 86.4\dev{4.0} & 99.9\dev{0.0} \\
Hazelnut & 96.9\dev{0.0} & 97.9\dev{0.6} & 93.3\dev{1.7} & 92.9\dev{2.2} & 98.6\dev{0.7} & 98.0\dev{1.1} & 94.1\dev{0.5} & 96.0\dev{2.0} & 99.1\dev{0.4} & 99.1\dev{0.7} & 94.8\dev{0.6} & 97.0\dev{1.2} & 99.1\dev{0.2} \\
Leather & 100.0\dev{0.0} & 100.0\dev{0.0} & 99.2\dev{0.2} & 99.1\dev{0.2} & 100.0\dev{0.0} & 100.0\dev{0.0} & 99.2\dev{0.3} & 99.3\dev{0.2} & 100.0\dev{0.0} & 100.0\dev{0.0} & 99.6\dev{0.1} & 99.6\dev{0.1} & 100.0\dev{0.0} \\
Metal nut & 99.3\dev{0.0} & 91.7\dev{0.8} & 82.0\dev{2.7} & 91.0\dev{1.1} & 99.7\dev{0.2} & 93.7\dev{2.4} & 82.2\dev{1.4} & 92.3\dev{4.0} & 99.9\dev{0.0} & 94.1\dev{1.8} & 85.5\dev{1.7} & 97.0\dev{2.6} & 99.9\dev{0.1} \\
Pill  & 95.7\dev{0.0} & 97.0\dev{0.8} & 88.3\dev{1.3} & 96.5\dev{0.6} & 98.3\dev{0.5} & 96.5\dev{0.4} & 87.9\dev{2.6} & 96.6\dev{0.7} & 98.6\dev{0.1} & 97.0\dev{0.2} & 87.0\dev{1.2} & 96.9\dev{0.4} & 98.6\dev{0.2} \\
Screw & 93.1\dev{0.0} & 71.3\dev{1.8} & 78.1\dev{1.0} & 71.4\dev{2.3} & 94.2\dev{0.6} & 71.0\dev{1.4} & 77.3\dev{1.3} & 72.9\dev{3.4} & 94.1\dev{1.5} & 73.7\dev{2.4} & 75.7\dev{2.8} & 71.8\dev{1.9} & 94.9\dev{0.8} \\
Tile  & 100.0\dev{0.0} & 100.0\dev{0.0} & 97.2\dev{0.7} & 99.6\dev{0.3} & 100.0\dev{0.0} & 100.0\dev{0.0} & 97.6\dev{0.4} & 99.4\dev{0.4} & 100.0\dev{0.1} & 100.0\dev{0.0} & 97.6\dev{0.2} & 99.6\dev{0.1} & 100.0\dev{0.0} \\
Toothbrush & 95.6\dev{0.0} & 88.3\dev{0.6} & 93.7\dev{0.5} & 93.5\dev{1.4} & 96.7\dev{2.0} & 90.8\dev{1.3} & 95.2\dev{1.6} & 94.1\dev{1.4} & 99.0\dev{0.6} & 91.3\dev{2.6} & 95.8\dev{0.7} & 94.8\dev{0.7} & 98.7\dev{1.1} \\
Transistor & 87.1\dev{0.0} & 76.2\dev{1.7} & 66.2\dev{7.5} & 77.7\dev{5.5} & 79.0\dev{4.0} & 81.6\dev{3.4} & 69.0\dev{6.5} & 89.3\dev{3.9} & 80.7\dev{2.3} & 80.3\dev{2.6} & 77.6\dev{8.4} & 84.5\dev{9.0} & 80.7\dev{3.2} \\
Wood  & 99.8\dev{0.0} & 99.6\dev{0.1} & 98.8\dev{0.3} & 99.3\dev{0.1} & 100.0\dev{0.0} & 99.7\dev{0.1} & 99.0\dev{0.1} & 99.5\dev{0.2} & 100.0\dev{0.0} & 99.7\dev{0.2} & 99.1\dev{0.0} & 99.5\dev{0.2} & 99.9\dev{0.1} \\
Zipper & 97.5\dev{0.0} & 96.9\dev{0.5} & 95.5\dev{0.9} & 97.2\dev{0.3} & 96.8\dev{1.8} & 98.2\dev{0.8} & 95.4\dev{1.0} & 97.8\dev{1.0} & 98.3\dev{0.4} & 98.6\dev{0.4} & 96.2\dev{0.8} & 99.1\dev{0.7} & 98.5\dev{0.2} \\
\cmidrule(r){1-1} \cmidrule(lr){2-2} \cmidrule(lr){3-6} \cmidrule(lr){7-10} \cmidrule(l){11-14}
Mean  & \textbf{96.5\dev{0.0}} & 90.6\dev{0.8} & 88.1\dev{1.7} & 92.2\dev{1.5} & \textbf{96.5\dev{0.9}} & 91.7\dev{1.2} & 89.3\dev{1.7} & 93.8\dev{1.7} & \textbf{97.0\dev{0.7}} & 92.5\dev{1.2} & 90.5\dev{1.6} & 94.5\dev{1.5} & \textbf{97.3\dev{0.6}} \\
\bottomrule
\end{tabular}%

  \end{adjustbox}
  \caption{Comparison of anomaly classification (AC) performance in terms of class-wise AUPR on MVTec-AD. We report the mean and standard deviation over 5 random seeds for each measurement.}
  \label{tab:mvtec/ac/aupr}
  \vspace{0.1in}
  \begin{adjustbox}{width=\linewidth}
\begin{tabular}{cccccccccccccc}
\toprule
MVTec-AD (AC) & $K=0$ & \multicolumn{4}{c}{$K=1$}     & \multicolumn{4}{c}{$K=2$}     & \multicolumn{4}{c}{$K=4$} \\
\cmidrule(lr){2-2} \cmidrule(lr){3-6} \cmidrule(lr){7-10} \cmidrule(l){11-14}
$F_1$-max & \textbf{WinCLIP} & SPADE & PaDiM & PatchCore & \textbf{WinCLIP+} & SPADE & PaDiM & PatchCore & \textbf{WinCLIP+} & SPADE & PaDiM & PatchCore & \textbf{WinCLIP+} \\
\cmidrule(r){1-1} \cmidrule(lr){2-2} \cmidrule(lr){3-6} \cmidrule(lr){7-10} \cmidrule(l){11-14}
Bottle & 97.6\dev{0.0} & 97.8\dev{0.8} & 96.3\dev{1.2} & 98.3\dev{0.6} & 96.5\dev{1.3} & 98.7\dev{0.4} & 97.1\dev{1.1} & 97.5\dev{0.6} & 97.7\dev{0.7} & 98.6\dev{0.3} & 97.9\dev{0.4} & 97.9\dev{0.8} & 97.8\dev{0.6} \\
Cable & 84.5\dev{0.0} & 79.6\dev{2.3} & 77.2\dev{1.1} & 85.2\dev{3.6} & 86.1\dev{1.3} & 80.4\dev{1.7} & 78.7\dev{1.2} & 86.1\dev{2.4} & 85.2\dev{0.7} & 83.8\dev{2.5} & 81.1\dev{1.1} & 91.3\dev{1.0} & 87.2\dev{0.6} \\
Capsule & 91.4\dev{0.0} & 92.0\dev{0.6} & 91.0\dev{0.2} & 92.0\dev{1.0} & 91.6\dev{0.7} & 92.1\dev{0.4} & 92.1\dev{0.9} & 93.6\dev{0.6} & 92.1\dev{0.7} & 92.7\dev{0.3} & 92.8\dev{0.9} & 94.3\dev{0.3} & 92.5\dev{0.5} \\
Carpet & 99.4\dev{0.0} & 96.5\dev{0.2} & 95.1\dev{0.5} & 94.9\dev{0.5} & 99.2\dev{0.8} & 96.6\dev{0.3} & 96.5\dev{0.4} & 95.3\dev{0.5} & 99.3\dev{0.7} & 96.9\dev{0.3} & 96.6\dev{0.3} & 94.3\dev{0.8} & 99.9\dev{0.2} \\
Grid  & 98.2\dev{0.0} & 84.5\dev{0.3} & 84.5\dev{0.3} & 86.2\dev{1.1} & 98.9\dev{0.4} & 84.8\dev{0.3} & 85.3\dev{0.9} & 86.9\dev{2.3} & 99.1\dev{0.0} & 84.8\dev{0.5} & 85.0\dev{0.5} & 87.5\dev{2.0} & 99.1\dev{0.0} \\
Hazelnut & 89.7\dev{0.0} & 92.4\dev{1.3} & 87.4\dev{1.6} & 87.0\dev{1.4} & 94.7\dev{2.3} & 93.2\dev{2.8} & 89.3\dev{1.1} & 91.0\dev{3.7} & 95.6\dev{1.6} & 95.9\dev{2.0} & 90.0\dev{1.7} & 92.8\dev{1.2} & 96.2\dev{1.0} \\
Leather & 100.0\dev{0.0} & 99.9\dev{0.2} & 96.2\dev{0.9} & 95.9\dev{0.7} & 99.5\dev{0.0} & 100.0\dev{0.0} & 96.6\dev{1.3} & 95.7\dev{0.9} & 99.7\dev{0.2} & 100.0\dev{0.0} & 97.9\dev{0.2} & 97.5\dev{0.7} & 99.8\dev{0.2} \\
Metal nut & 96.3\dev{0.0} & 90.1\dev{0.6} & 90.1\dev{0.3} & 91.4\dev{1.2} & 97.7\dev{1.0} & 90.5\dev{1.1} & 90.0\dev{0.3} & 91.9\dev{0.9} & 98.4\dev{0.5} & 90.6\dev{0.9} & 90.3\dev{0.4} & 93.6\dev{1.4} & 98.5\dev{0.6} \\
Pill  & 91.6\dev{0.0} & 93.5\dev{0.4} & 91.6\dev{0.0} & 91.9\dev{0.3} & 93.8\dev{0.7} & 93.4\dev{0.3} & 91.7\dev{0.1} & 92.0\dev{0.3} & 94.3\dev{0.4} & 93.6\dev{0.5} & 91.7\dev{0.1} & 92.1\dev{0.3} & 94.1\dev{0.4} \\
Screw & 87.4\dev{0.0} & 85.3\dev{0.0} & 85.6\dev{0.3} & 85.8\dev{0.2} & 88.5\dev{0.3} & 85.6\dev{0.2} & 85.5\dev{0.1} & 85.7\dev{0.2} & 89.0\dev{0.6} & 85.8\dev{0.7} & 85.7\dev{0.3} & 86.8\dev{0.6} & 89.6\dev{0.7} \\
Tile  & 99.4\dev{0.0} & 99.2\dev{0.5} & 90.7\dev{2.0} & 97.5\dev{1.2} & 98.9\dev{0.2} & 99.2\dev{0.3} & 91.5\dev{1.3} & 96.9\dev{1.5} & 99.2\dev{0.3} & 99.4\dev{0.0} & 91.0\dev{0.9} & 97.5\dev{0.4} & 99.2\dev{0.3} \\
Toothbrush & 87.9\dev{0.0} & 85.6\dev{1.3} & 85.4\dev{1.3} & 88.9\dev{2.2} & 94.1\dev{1.9} & 87.1\dev{1.4} & 90.1\dev{3.1} & 90.8\dev{1.6} & 96.7\dev{1.8} & 86.5\dev{1.8} & 90.6\dev{2.1} & 92.6\dev{2.2} & 96.8\dev{2.3} \\
Transistor & 79.5\dev{0.0} & 70.3\dev{1.7} & 66.4\dev{4.7} & 70.6\dev{7.2} & 75.1\dev{3.1} & 73.4\dev{3.4} & 65.9\dev{3.7} & 85.3\dev{6.1} & 75.9\dev{2.4} & 72.3\dev{2.7} & 74.8\dev{7.7} & 78.3\dev{11.5} & 76.6\dev{2.8} \\
Wood  & 98.3\dev{0.0} & 97.0\dev{0.8} & 94.3\dev{1.1} & 96.1\dev{0.3} & 99.4\dev{0.3} & 97.7\dev{1.0} & 94.9\dev{0.3} & 96.6\dev{0.9} & 99.5\dev{0.4} & 97.6\dev{0.7} & 94.8\dev{0.5} & 96.5\dev{0.9} & 99.2\dev{0.9} \\
Zipper & 92.9\dev{0.0} & 91.0\dev{0.9} & 91.2\dev{0.8} & 95.3\dev{0.5} & 92.1\dev{2.5} & 93.5\dev{1.8} & 92.2\dev{0.6} & 95.4\dev{0.5} & 94.4\dev{0.3} & 94.7\dev{0.8} & 92.5\dev{0.8} & 96.5\dev{0.3} & 94.7\dev{0.4} \\
\cmidrule(r){1-1} \cmidrule(lr){2-2} \cmidrule(lr){3-6} \cmidrule(lr){7-10} \cmidrule(l){11-14}
Mean  & \textbf{92.9\dev{0.0}} & 90.3\dev{0.8} & 88.2\dev{1.1} & 90.5\dev{1.5} & \textbf{93.7\dev{1.1}} & 91.1\dev{1.0} & 89.2\dev{1.1} & 92.0\dev{1.5} & \textbf{94.4\dev{0.8}} & 91.5\dev{0.9} & 90.2\dev{1.2} & 92.6\dev{1.6} & \textbf{94.7\dev{0.8}} \\
\bottomrule
\end{tabular}%

  \end{adjustbox}
  \caption{Comparison of anomaly classification (AC) performance in terms of class-wise $F_1$-max on MVTec-AD. We report the mean and standard deviation over 5 random seeds for each measurement.}
  \label{tab:mvtec/ac/f1}
\end{table*}

\begin{table*}[!ht]
  \centering
  \begin{adjustbox}{width=\linewidth}
\begin{tabular}{cccccccccccccc}
\toprule
MVTec-AD (AS) & $K=0$ & \multicolumn{4}{c}{$K=1$}     & \multicolumn{4}{c}{$K=2$}     & \multicolumn{4}{c}{$K=4$} \\
\cmidrule(lr){2-2} \cmidrule(lr){3-6} \cmidrule(lr){7-10} \cmidrule(l){11-14}
pAUROC & \textbf{WinCLIP} & SPADE & PaDiM & PatchCore & \textbf{WinCLIP+} & SPADE & PaDiM & PatchCore & \textbf{WinCLIP+} & SPADE & PaDiM & PatchCore & \textbf{WinCLIP+} \\
\cmidrule(r){1-1} \cmidrule(lr){2-2} \cmidrule(lr){3-6} \cmidrule(lr){7-10} \cmidrule(l){11-14}
Bottle & 89.5\dev{0.0} & 95.3\dev{0.2} & 96.1\dev{0.5} & 97.9\dev{0.1} & 97.5\dev{0.2} & 95.7\dev{0.2} & 96.9\dev{0.1} & 98.1\dev{0.0} & 97.7\dev{0.1} & 96.1\dev{0.0} & 97.1\dev{0.1} & 98.2\dev{0.0} & 97.8\dev{0.0} \\
Cable & 77.0\dev{0.0} & 86.4\dev{0.2} & 88.4\dev{1.2} & 95.5\dev{0.8} & 93.8\dev{0.6} & 87.4\dev{0.3} & 90.0\dev{0.8} & 96.4\dev{0.3} & 94.3\dev{0.4} & 88.2\dev{0.2} & 92.1\dev{0.4} & 97.5\dev{0.3} & 94.9\dev{0.1} \\
Capsule & 86.9\dev{0.0} & 96.3\dev{0.2} & 94.5\dev{0.6} & 95.6\dev{0.4} & 94.6\dev{0.8} & 96.7\dev{0.1} & 95.2\dev{0.5} & 96.5\dev{0.4} & 96.4\dev{0.3} & 97.0\dev{0.2} & 96.2\dev{0.4} & 96.8\dev{0.6} & 96.2\dev{0.5} \\
Carpet & 95.4\dev{0.0} & 98.2\dev{0.0} & 97.8\dev{0.2} & 98.4\dev{0.1} & 99.4\dev{0.0} & 98.3\dev{0.0} & 98.2\dev{0.0} & 98.5\dev{0.1} & 99.3\dev{0.0} & 98.4\dev{0.0} & 98.4\dev{0.0} & 98.6\dev{0.1} & 99.3\dev{0.0} \\
Grid  & 82.2\dev{0.0} & 80.7\dev{1.3} & 70.2\dev{2.8} & 58.8\dev{4.9} & 96.8\dev{1.0} & 83.5\dev{1.0} & 70.8\dev{2.0} & 62.6\dev{3.2} & 97.7\dev{0.8} & 87.2\dev{1.1} & 77.0\dev{1.8} & 69.4\dev{1.3} & 98.0\dev{0.2} \\
Hazelnut & 94.3\dev{0.0} & 97.2\dev{0.1} & 95.4\dev{0.7} & 95.8\dev{0.6} & 98.5\dev{0.2} & 97.6\dev{0.1} & 96.8\dev{0.3} & 96.3\dev{0.6} & 98.7\dev{0.1} & 97.7\dev{0.1} & 97.2\dev{0.2} & 97.6\dev{0.1} & 98.8\dev{0.0} \\
Leather & 96.7\dev{0.0} & 99.1\dev{0.0} & 98.5\dev{0.1} & 98.8\dev{0.2} & 99.3\dev{0.0} & 99.1\dev{0.0} & 98.7\dev{0.1} & 99.0\dev{0.1} & 99.3\dev{0.0} & 99.1\dev{0.0} & 98.8\dev{0.0} & 99.1\dev{0.0} & 99.3\dev{0.0} \\
Metal nut & 61.0\dev{0.0} & 83.8\dev{0.7} & 74.6\dev{1.1} & 89.3\dev{1.4} & 90.0\dev{0.6} & 85.8\dev{1.1} & 80.3\dev{2.1} & 94.6\dev{1.4} & 91.4\dev{0.4} & 87.1\dev{0.7} & 82.7\dev{3.9} & 95.9\dev{1.8} & 92.9\dev{0.4} \\
Pill  & 80.0\dev{0.0} & 89.4\dev{0.4} & 84.8\dev{1.0} & 93.1\dev{1.1} & 96.4\dev{0.3} & 89.9\dev{0.2} & 87.3\dev{0.7} & 94.2\dev{0.3} & 97.0\dev{0.2} & 90.7\dev{0.2} & 88.9\dev{0.5} & 94.8\dev{0.4} & 97.1\dev{0.0} \\
Screw & 89.6\dev{0.0} & 94.8\dev{0.2} & 83.3\dev{0.7} & 89.6\dev{0.5} & 94.5\dev{0.4} & 95.6\dev{0.4} & 89.8\dev{0.8} & 90.0\dev{0.7} & 95.2\dev{0.3} & 96.4\dev{0.4} & 90.8\dev{0.2} & 91.3\dev{1.0} & 96.0\dev{0.5} \\
Tile  & 77.6\dev{0.0} & 91.7\dev{0.3} & 84.1\dev{1.1} & 94.1\dev{0.5} & 96.3\dev{0.2} & 92.0\dev{0.1} & 87.7\dev{0.2} & 94.4\dev{0.2} & 96.5\dev{0.1} & 92.2\dev{0.1} & 88.9\dev{0.3} & 94.6\dev{0.1} & 96.6\dev{0.1} \\
Toothbrush & 86.9\dev{0.0} & 94.6\dev{0.6} & 97.3\dev{0.3} & 97.3\dev{0.4} & 97.8\dev{0.1} & 96.2\dev{0.3} & 97.7\dev{0.3} & 97.5\dev{0.2} & 98.1\dev{0.1} & 97.0\dev{0.6} & 98.4\dev{0.2} & 98.4\dev{0.4} & 98.4\dev{0.5} \\
Transistor & 74.7\dev{0.0} & 71.4\dev{1.3} & 90.2\dev{2.8} & 84.9\dev{2.7} & 85.0\dev{1.8} & 72.8\dev{0.9} & 92.3\dev{2.1} & 89.6\dev{0.9} & 88.3\dev{1.0} & 73.4\dev{0.7} & 94.0\dev{2.7} & 90.7\dev{1.4} & 88.5\dev{1.2} \\
Wood  & 93.4\dev{0.0} & 93.4\dev{0.1} & 90.7\dev{0.4} & 92.7\dev{0.9} & 94.6\dev{1.0} & 93.8\dev{0.1} & 91.9\dev{0.1} & 93.2\dev{0.7} & 95.3\dev{0.4} & 93.9\dev{0.1} & 92.2\dev{0.1} & 93.5\dev{0.3} & 95.4\dev{0.2} \\
Zipper & 91.6\dev{0.0} & 94.9\dev{0.3} & 93.9\dev{0.8} & 97.4\dev{0.4} & 93.9\dev{0.8} & 95.8\dev{0.2} & 95.4\dev{0.3} & 98.0\dev{0.1} & 94.1\dev{0.7} & 96.2\dev{0.1} & 96.1\dev{0.2} & 98.1\dev{0.1} & 94.2\dev{0.4} \\
\cmidrule(r){1-1} \cmidrule(lr){2-2} \cmidrule(lr){3-6} \cmidrule(lr){7-10} \cmidrule(l){11-14}
Mean  & \textbf{85.1\dev{0.0}} & 91.2\dev{0.4} & 89.3\dev{0.9} & 92.0\dev{1.0} & \textbf{95.2\dev{0.5}} & 92.0\dev{0.3} & 91.3\dev{0.7} & 93.3\dev{0.6} & \textbf{96.0\dev{0.3}} & 92.7\dev{0.3} & 92.6\dev{0.7} & 94.3\dev{0.5} & \textbf{96.2\dev{0.3}} \\
\bottomrule
\end{tabular}%

  \end{adjustbox}
  \caption{Comparison of anomaly segmentation (AS) performance in terms of class-wise pixel-AUROC on MVTec-AD. We report the mean and standard deviation over 5 random seeds for each measurement.}
  \label{tab:mvtec/as/proc}
  \vspace{0.1in}
  \begin{adjustbox}{width=\linewidth}
\begin{tabular}{cccccccccccccc}
\toprule
MVTec-AD (AS) & $K=0$ & \multicolumn{4}{c}{$K=1$}     & \multicolumn{4}{c}{$K=2$}     & \multicolumn{4}{c}{$K=4$} \\
\cmidrule(lr){2-2} \cmidrule(lr){3-6} \cmidrule(lr){7-10} \cmidrule(l){11-14}
PRO   & \textbf{WinCLIP} & SPADE & PaDiM & PatchCore & \textbf{WinCLIP+} & SPADE & PaDiM & PatchCore & \textbf{WinCLIP+} & SPADE & PaDiM & PatchCore & \textbf{WinCLIP+} \\
\cmidrule(r){1-1} \cmidrule(lr){2-2} \cmidrule(lr){3-6} \cmidrule(lr){7-10} \cmidrule(l){11-14}
Bottle & 76.4\dev{0.0} & 91.1\dev{0.4} & 89.8\dev{0.8} & 93.5\dev{0.3} & 91.2\dev{0.4} & 91.8\dev{0.5} & 91.7\dev{0.2} & 93.9\dev{0.3} & 91.8\dev{0.3} & 92.5\dev{0.1} & 92.2\dev{0.2} & 94.0\dev{0.2} & 91.6\dev{0.2} \\
Cable & 42.9\dev{0.0} & 63.5\dev{0.7} & 59.1\dev{3.2} & 84.7\dev{1.0} & 72.5\dev{2.3} & 66.7\dev{0.9} & 66.5\dev{2.8} & 88.5\dev{0.9} & 74.7\dev{2.3} & 69.5\dev{0.4} & 74.2\dev{1.8} & 91.7\dev{0.6} & 77.0\dev{1.1} \\
Capsule & 62.1\dev{0.0} & 92.7\dev{0.4} & 80.0\dev{2.0} & 83.9\dev{0.9} & 85.6\dev{2.7} & 93.4\dev{0.3} & 82.3\dev{2.1} & 86.6\dev{1.0} & 90.6\dev{0.6} & 94.1\dev{0.6} & 85.7\dev{1.3} & 87.8\dev{1.9} & 90.1\dev{1.5} \\
Carpet & 84.1\dev{0.0} & 96.1\dev{0.0} & 92.9\dev{0.3} & 93.3\dev{0.3} & 97.4\dev{0.4} & 96.2\dev{0.0} & 93.9\dev{0.2} & 93.7\dev{0.4} & 97.3\dev{0.3} & 96.3\dev{0.0} & 94.4\dev{0.2} & 93.9\dev{0.4} & 97.0\dev{0.2} \\
Grid  & 57.0\dev{0.0} & 67.7\dev{1.9} & 41.2\dev{4.6} & 21.7\dev{9.5} & 90.5\dev{2.7} & 72.1\dev{1.5} & 45.1\dev{3.6} & 23.7\dev{3.8} & 92.8\dev{2.5} & 78.0\dev{1.5} & 55.5\dev{3.4} & 30.4\dev{4.6} & 93.6\dev{0.6} \\
Hazelnut & 81.6\dev{0.0} & 94.9\dev{0.3} & 85.7\dev{1.9} & 88.3\dev{1.3} & 93.7\dev{0.9} & 95.6\dev{0.2} & 89.4\dev{0.9} & 89.8\dev{1.3} & 94.2\dev{0.3} & 95.6\dev{0.1} & 90.4\dev{0.7} & 92.0\dev{0.3} & 94.2\dev{0.3} \\
Leather & 91.1\dev{0.0} & 98.7\dev{0.0} & 95.6\dev{0.2} & 95.2\dev{1.0} & 98.6\dev{0.0} & 98.8\dev{0.0} & 96.2\dev{0.2} & 95.9\dev{0.3} & 98.3\dev{0.4} & 98.8\dev{0.0} & 96.3\dev{0.1} & 96.4\dev{0.1} & 98.0\dev{0.4} \\
Metal nut & 31.8\dev{0.0} & 73.4\dev{1.1} & 38.1\dev{1.6} & 66.7\dev{2.9} & 84.7\dev{1.1} & 78.1\dev{1.8} & 48.2\dev{5.0} & 79.6\dev{4.2} & 86.7\dev{0.8} & 81.2\dev{1.4} & 54.0\dev{8.8} & 83.8\dev{5.5} & 89.4\dev{0.1} \\
Pill  & 65.0\dev{0.0} & 92.8\dev{0.3} & 78.9\dev{0.6} & 89.5\dev{1.6} & 93.5\dev{0.2} & 93.3\dev{0.2} & 84.3\dev{0.4} & 91.6\dev{0.5} & 94.5\dev{0.2} & 93.9\dev{0.2} & 86.6\dev{0.4} & 92.5\dev{0.4} & 94.6\dev{0.3} \\
Screw & 68.5\dev{0.0} & 85.0\dev{0.8} & 51.6\dev{1.7} & 68.1\dev{1.3} & 82.3\dev{1.1} & 87.2\dev{1.2} & 69.5\dev{2.1} & 69.0\dev{2.1} & 84.1\dev{0.5} & 89.5\dev{1.3} & 72.3\dev{0.8} & 72.4\dev{3.1} & 86.3\dev{1.8} \\
Tile  & 51.2\dev{0.0} & 84.2\dev{0.4} & 66.7\dev{1.5} & 82.5\dev{1.1} & 89.4\dev{0.4} & 84.6\dev{0.2} & 71.9\dev{0.5} & 82.5\dev{0.5} & 89.6\dev{0.4} & 84.9\dev{0.1} & 73.6\dev{0.9} & 83.0\dev{0.1} & 89.9\dev{0.3} \\
Toothbrush & 67.7\dev{0.0} & 83.5\dev{1.3} & 82.1\dev{1.5} & 79.0\dev{2.4} & 85.3\dev{1.0} & 87.4\dev{1.1} & 83.3\dev{2.6} & 81.0\dev{0.7} & 84.7\dev{1.4} & 89.0\dev{1.1} & 87.1\dev{1.7} & 85.5\dev{3.0} & 86.0\dev{3.3} \\
Transistor & 43.4\dev{0.0} & 55.3\dev{2.0} & 70.3\dev{7.0} & 70.9\dev{4.6} & 65.0\dev{1.8} & 57.6\dev{1.4} & 76.5\dev{5.5} & 78.8\dev{1.5} & 68.6\dev{1.1} & 58.5\dev{0.7} & 82.2\dev{7.4} & 79.5\dev{2.8} & 69.0\dev{1.1} \\
Wood  & 74.1\dev{0.0} & 92.9\dev{0.1} & 86.5\dev{0.6} & 87.1\dev{1.0} & 91.0\dev{0.6} & 93.1\dev{0.1} & 88.0\dev{0.2} & 86.8\dev{1.4} & 91.8\dev{0.6} & 93.2\dev{0.1} & 88.4\dev{0.2} & 87.7\dev{0.4} & 91.7\dev{0.3} \\
Zipper & 71.7\dev{0.0} & 86.8\dev{0.6} & 81.7\dev{2.0} & 91.2\dev{1.1} & 86.0\dev{1.7} & 89.0\dev{0.4} & 85.6\dev{0.7} & 92.8\dev{0.4} & 86.4\dev{1.6} & 90.1\dev{0.2} & 87.2\dev{0.8} & 93.4\dev{0.2} & 86.9\dev{0.7} \\
\cmidrule(r){1-1} \cmidrule(lr){2-2} \cmidrule(lr){3-6} \cmidrule(lr){7-10} \cmidrule(l){11-14}
Mean  & \textbf{64.6\dev{0.0}} & 83.9\dev{0.7} & 73.3\dev{2.0} & 79.7\dev{2.0} & \textbf{87.1\dev{1.2}} & 85.7\dev{0.7} & 78.2\dev{1.8} & 82.3\dev{1.3} & \textbf{88.4\dev{0.9}} & 87.0\dev{0.5} & 81.3\dev{1.9} & 84.3\dev{1.6} & \textbf{89.0\dev{0.8}} \\
\bottomrule
\end{tabular}%

  \end{adjustbox}
  \caption{Comparison of anomaly segmentation (AS) performance in terms of class-wise PRO on MVTec-AD. We report the mean and standard deviation over 5 random seeds for each measurement.}
  \label{tab:mvtec/as/pro}
  \vspace{0.1in}
  \begin{adjustbox}{width=\linewidth}
\begin{tabular}{cccccccccccccc}
\toprule
MVTec-AD (AS) & $K=0$ & \multicolumn{4}{c}{$K=1$}     & \multicolumn{4}{c}{$K=2$}     & \multicolumn{4}{c}{$K=4$} \\
\cmidrule(lr){2-2} \cmidrule(lr){3-6} \cmidrule(lr){7-10} \cmidrule(l){11-14}
$F_1$-max & \textbf{WinCLIP} & SPADE & PaDiM & PatchCore & \textbf{WinCLIP+} & SPADE & PaDiM & PatchCore & \textbf{WinCLIP+} & SPADE & PaDiM & PatchCore & \textbf{WinCLIP+} \\
\cmidrule(r){1-1} \cmidrule(lr){2-2} \cmidrule(lr){3-6} \cmidrule(lr){7-10} \cmidrule(l){11-14}
Bottle & 58.1\dev{0.0} & 61.5\dev{0.3} & 68.2\dev{1.9} & 74.8\dev{0.4} & 72.8\dev{0.8} & 62.7\dev{0.4} & 70.7\dev{0.4} & 75.1\dev{0.1} & 73.2\dev{0.9} & 64.3\dev{0.3} & 71.4\dev{0.4} & 75.0\dev{0.2} & 73.3\dev{0.6} \\
Cable & 19.7\dev{0.0} & 25.9\dev{1.2} & 27.4\dev{1.8} & 59.8\dev{1.4} & 49.4\dev{3.3} & 28.5\dev{0.8} & 29.5\dev{1.6} & 62.2\dev{1.0} & 51.2\dev{1.3} & 30.2\dev{0.4} & 34.5\dev{1.1} & 65.5\dev{1.1} & 54.7\dev{1.1} \\
Capsule & 21.7\dev{0.0} & 37.5\dev{3.5} & 27.1\dev{2.8} & 32.3\dev{2.1} & 29.7\dev{7.8} & 39.6\dev{3.0} & 33.1\dev{2.6} & 37.9\dev{4.5} & 43.5\dev{1.4} & 40.8\dev{3.4} & 37.0\dev{2.0} & 39.0\dev{6.3} & 40.7\dev{4.9} \\
Carpet & 49.7\dev{0.0} & 67.1\dev{0.2} & 62.4\dev{0.5} & 67.3\dev{0.4} & 73.3\dev{1.5} & 67.6\dev{0.2} & 62.6\dev{0.2} & 67.0\dev{0.7} & 72.9\dev{1.3} & 68.1\dev{0.2} & 62.9\dev{0.2} & 67.4\dev{0.3} & 72.0\dev{0.7} \\
Grid  & 18.6\dev{0.0} & 17.0\dev{1.0} & 9.4\dev{2.1} & 5.5\dev{2.2} & 50.7\dev{4.5} & 18.9\dev{1.1} & 13.1\dev{1.5} & 5.2\dev{1.2} & 53.4\dev{3.8} & 23.1\dev{1.6} & 18.0\dev{1.9} & 10.0\dev{5.3} & 52.7\dev{1.5} \\
Hazelnut & 37.6\dev{0.0} & 62.8\dev{0.8} & 47.7\dev{3.3} & 50.1\dev{3.9} & 68.9\dev{2.6} & 65.1\dev{0.3} & 57.1\dev{0.7} & 53.6\dev{3.7} & 70.5\dev{1.7} & 65.2\dev{0.6} & 58.0\dev{1.3} & 60.8\dev{1.5} & 71.0\dev{0.3} \\
Leather & 39.7\dev{0.0} & 55.6\dev{0.1} & 52.3\dev{0.8} & 58.6\dev{0.4} & 58.0\dev{0.7} & 55.8\dev{0.6} & 52.8\dev{0.2} & 58.8\dev{0.3} & 57.5\dev{0.6} & 55.5\dev{0.1} & 52.5\dev{0.2} & 58.8\dev{0.3} & 56.3\dev{1.0} \\
Metal nut & 32.4\dev{0.0} & 46.4\dev{1.1} & 38.2\dev{0.9} & 55.1\dev{2.6} & 59.4\dev{1.7} & 48.7\dev{1.4} & 44.5\dev{2.0} & 70.4\dev{4.8} & 62.7\dev{1.5} & 50.4\dev{0.9} & 47.5\dev{4.4} & 74.8\dev{6.7} & 67.4\dev{1.6} \\
Pill  & 17.6\dev{0.0} & 29.6\dev{0.8} & 25.3\dev{0.6} & 54.5\dev{4.0} & 64.7\dev{1.8} & 31.2\dev{0.4} & 28.8\dev{1.2} & 59.4\dev{1.7} & 67.8\dev{0.5} & 33.0\dev{0.5} & 32.7\dev{1.1} & 61.7\dev{1.6} & 67.9\dev{0.4} \\
Screw & 13.5\dev{0.0} & 11.6\dev{1.0} & 3.5\dev{0.1} & 6.4\dev{0.4} & 22.2\dev{2.8} & 14.7\dev{2.3} & 5.9\dev{0.3} & 6.5\dev{0.4} & 22.4\dev{2.8} & 20.1\dev{5.2} & 6.4\dev{0.2} & 7.4\dev{0.5} & 30.1\dev{4.3} \\
Tile  & 32.6\dev{0.0} & 57.3\dev{0.5} & 42.2\dev{1.4} & 64.2\dev{1.4} & 71.2\dev{0.4} & 58.0\dev{0.2} & 47.3\dev{0.4} & 64.4\dev{0.8} & 71.9\dev{0.6} & 58.4\dev{0.2} & 48.8\dev{0.6} & 65.0\dev{0.1} & 72.2\dev{0.6} \\
Toothbrush & 17.1\dev{0.0} & 40.0\dev{2.5} & 59.5\dev{3.8} & 63.4\dev{3.2} & 62.7\dev{3.6} & 46.9\dev{1.8} & 62.4\dev{2.7} & 61.5\dev{2.4} & 65.8\dev{2.2} & 51.0\dev{3.7} & 65.0\dev{1.4} & 64.9\dev{0.5} & 69.4\dev{4.6} \\
Transistor & 30.5\dev{0.0} & 21.4\dev{1.9} & 41.6\dev{8.1} & 48.2\dev{5.6} & 39.1\dev{3.5} & 23.2\dev{1.4} & 47.5\dev{7.7} & 54.6\dev{1.5} & 45.6\dev{2.3} & 23.8\dev{0.8} & 54.0\dev{10.5} & 55.7\dev{2.6} & 46.6\dev{2.2} \\
Wood  & 51.5\dev{0.0} & 56.4\dev{0.3} & 46.6\dev{0.7} & 52.9\dev{1.1} & 65.2\dev{1.4} & 57.0\dev{0.1} & 47.2\dev{0.2} & 52.9\dev{1.8} & 65.8\dev{0.6} & 57.1\dev{0.4} & 47.7\dev{0.3} & 53.3\dev{0.7} & 65.1\dev{0.5} \\
Zipper & 34.4\dev{0.0} & 45.1\dev{0.3} & 51.3\dev{2.5} & 62.5\dev{2.4} & 50.6\dev{3.9} & 48.7\dev{0.5} & 53.1\dev{1.1} & 65.3\dev{0.7} & 50.9\dev{4.5} & 51.2\dev{0.6} & 55.2\dev{1.7} & 65.1\dev{0.5} & 52.8\dev{2.7} \\
\cmidrule(r){1-1} \cmidrule(lr){2-2} \cmidrule(lr){3-6} \cmidrule(lr){7-10} \cmidrule(l){11-14}
Mean  & \textbf{31.7\dev{0.0}} & 42.4\dev{1.0} & 40.2\dev{2.1} & 50.4\dev{2.1} & \textbf{55.9\dev{2.7}} & 44.5\dev{1.0} & 43.7\dev{1.5} & 53.0\dev{1.7} & \textbf{58.4\dev{1.7}} & 46.2\dev{1.3} & 46.1\dev{1.8} & 55.0\dev{1.9} & \textbf{59.5\dev{1.8}} \\
\bottomrule
\end{tabular}%

  \end{adjustbox}
  \caption{Comparison of anomaly segmentation (AS) performance in terms of class-wise $F_1$-max on MVTec-AD. We report the mean and standard deviation over 5 random seeds for each measurement.}
  \label{tab:mvtec/as/pf1}
\end{table*}

\begin{table*}[!ht]
  \centering
  \begin{adjustbox}{width=\linewidth}
\begin{tabular}{cccccccccccccc}
\toprule
VisA (AC) & $K=0$ & \multicolumn{4}{c}{$K=1$}     & \multicolumn{4}{c}{$K=2$}     & \multicolumn{4}{c}{$K=4$} \\
\cmidrule(lr){2-2} \cmidrule(lr){3-6} \cmidrule(lr){7-10} \cmidrule(l){11-14}
AUROC & \textbf{WinCLIP} & SPADE & PaDiM & PatchCore & \textbf{WinCLIP+} & SPADE & PaDiM & PatchCore & \textbf{WinCLIP+} & SPADE & PaDiM & PatchCore & \textbf{WinCLIP+} \\
\cmidrule(r){1-1} \cmidrule(lr){2-2} \cmidrule(lr){3-6} \cmidrule(lr){7-10} \cmidrule(l){11-14}
Candle & 95.4\dev{0.0} & 86.1\dev{5.6} & 70.8\dev{4.1} & 85.1\dev{1.4} & 93.4\dev{1.4} & 91.3\dev{3.3} & 75.8\dev{2.1} & 85.3\dev{1.5} & 94.8\dev{1.0} & 92.8\dev{2.1} & 77.5\dev{1.6} & 87.8\dev{0.8} & 95.1\dev{0.3} \\
Capsules & 85.0\dev{0.0} & 73.3\dev{7.5} & 51.0\dev{7.8} & 60.0\dev{7.6} & 85.0\dev{3.1} & 71.7\dev{11.2} & 51.7\dev{4.6} & 57.8\dev{5.4} & 84.9\dev{0.8} & 73.4\dev{7.1} & 52.7\dev{3.4} & 63.4\dev{5.4} & 86.8\dev{1.7} \\
Cashew & 92.1\dev{0.0} & 95.9\dev{1.1} & 62.3\dev{9.9} & 89.5\dev{4.4} & 94.0\dev{0.4} & 97.3\dev{1.4} & 74.6\dev{3.6} & 93.6\dev{0.6} & 94.3\dev{0.5} & 96.4\dev{1.3} & 77.7\dev{3.2} & 93.0\dev{1.5} & 95.2\dev{0.8} \\
Chewinggum & 96.5\dev{0.0} & 92.1\dev{2.0} & 69.9\dev{4.9} & 97.3\dev{0.3} & 97.6\dev{0.8} & 93.4\dev{1.0} & 82.7\dev{2.1} & 97.8\dev{0.6} & 97.3\dev{0.8} & 93.5\dev{1.4} & 83.5\dev{3.7} & 98.3\dev{0.3} & 97.7\dev{0.3} \\
Fryum & 80.3\dev{0.0} & 81.1\dev{4.0} & 58.3\dev{5.9} & 75.0\dev{4.8} & 88.5\dev{1.9} & 90.5\dev{3.9} & 69.2\dev{9.0} & 83.4\dev{2.4} & 90.5\dev{0.4} & 92.9\dev{1.6} & 71.2\dev{5.9} & 88.6\dev{1.3} & 90.8\dev{0.5} \\
Macaroni1 & 76.2\dev{0.0} & 66.0\dev{10.5} & 62.1\dev{4.6} & 68.0\dev{3.4} & 82.9\dev{1.5} & 69.1\dev{8.2} & 62.2\dev{5.0} & 75.6\dev{4.6} & 83.3\dev{1.9} & 65.8\dev{1.2} & 65.9\dev{3.9} & 82.9\dev{2.7} & 85.2\dev{0.9} \\
Macaroni2 & 63.7\dev{0.0} & 55.8\dev{6.1} & 47.5\dev{5.9} & 55.6\dev{4.6} & 70.2\dev{0.9} & 58.3\dev{4.4} & 50.8\dev{2.9} & 57.3\dev{5.6} & 71.8\dev{2.0} & 56.7\dev{3.2} & 55.0\dev{2.9} & 61.7\dev{1.8} & 70.9\dev{2.2} \\
PCB1  & 73.6\dev{0.0} & 87.2\dev{2.3} & 76.2\dev{1.2} & 78.9\dev{1.1} & 75.6\dev{23.0} & 86.7\dev{1.1} & 62.4\dev{10.8} & 71.5\dev{20.0} & 76.7\dev{5.2} & 83.4\dev{8.5} & 82.6\dev{1.5} & 84.7\dev{6.7} & 88.3\dev{1.7} \\
PCB2  & 51.2\dev{0.0} & 73.5\dev{3.7} & 61.2\dev{2.0} & 81.5\dev{0.8} & 62.2\dev{3.9} & 70.3\dev{8.1} & 66.8\dev{2.0} & 84.3\dev{1.7} & 62.6\dev{3.7} & 71.7\dev{7.0} & 73.5\dev{2.4} & 84.3\dev{1.0} & 67.5\dev{2.6} \\
PCB3  & 73.4\dev{0.0} & 72.2\dev{1.0} & 51.4\dev{12.2} & 82.7\dev{2.3} & 74.1\dev{1.1} & 75.8\dev{5.7} & 67.3\dev{3.8} & 84.8\dev{1.2} & 78.8\dev{1.9} & 79.0\dev{4.1} & 65.9\dev{1.9} & 87.0\dev{1.1} & 83.3\dev{1.7} \\
PCB4  & 79.6\dev{0.0} & 93.4\dev{1.3} & 76.1\dev{3.6} & 93.9\dev{2.8} & 85.2\dev{8.9} & 86.1\dev{8.2} & 69.3\dev{13.7} & 94.3\dev{3.2} & 82.3\dev{9.9} & 95.4\dev{2.3} & 85.4\dev{2.0} & 95.6\dev{1.6} & 87.6\dev{8.0} \\
Pipe fryum & 69.7\dev{0.0} & 77.9\dev{3.2} & 66.7\dev{2.2} & 90.7\dev{1.7} & 97.2\dev{1.1} & 78.1\dev{3.0} & 75.3\dev{1.8} & 93.5\dev{1.3} & 98.0\dev{0.6} & 79.3\dev{0.9} & 82.9\dev{2.2} & 96.4\dev{0.7} & 98.5\dev{0.4} \\
\cmidrule(r){1-1} \cmidrule(lr){2-2} \cmidrule(lr){3-6} \cmidrule(lr){7-10} \cmidrule(l){11-14}
Mean  & \textbf{78.1\dev{0.0}} & 79.5\dev{4.0} & 62.8\dev{5.4} & 79.9\dev{2.9} & \textbf{83.8\dev{4.0}} & 80.7\dev{5.0} & 67.4\dev{5.1} & 81.6\dev{4.0} & \textbf{84.6\dev{2.4}} & 81.7\dev{3.4} & 72.8\dev{2.9} & 85.3\dev{2.1} & \textbf{87.3\dev{1.8}} \\
\bottomrule
\end{tabular}%

  \end{adjustbox}
  \caption{Comparison of anomaly classification (AC) performance in terms of class-wise AUROC on VisA. We report the mean and standard deviation over 5 random seeds for each measurement.}
  \label{tab:visa/ac/roc}
  \vspace{0.1in}
  \begin{adjustbox}{width=\linewidth}
\begin{tabular}{cccccccccccccc}
\toprule
VisA (AC) & $K=0$ & \multicolumn{4}{c}{$K=1$}     & \multicolumn{4}{c}{$K=2$}     & \multicolumn{4}{c}{$K=4$} \\
\cmidrule(lr){2-2} \cmidrule(lr){3-6} \cmidrule(lr){7-10} \cmidrule(l){11-14}
AUPR  & \textbf{WinCLIP} & SPADE & PaDiM & PatchCore & \textbf{WinCLIP+} & SPADE & PaDiM & PatchCore & \textbf{WinCLIP+} & SPADE & PaDiM & PatchCore & \textbf{WinCLIP+} \\
\cmidrule(r){1-1} \cmidrule(lr){2-2} \cmidrule(lr){3-6} \cmidrule(lr){7-10} \cmidrule(l){11-14}
Candle & 95.8\dev{0.0} & 86.5\dev{4.3} & 69.2\dev{3.9} & 86.6\dev{2.3} & 93.6\dev{1.5} & 90.7\dev{3.2} & 72.8\dev{1.0} & 86.8\dev{1.7} & 95.1\dev{1.1} & 92.6\dev{1.9} & 72.5\dev{1.1} & 88.9\dev{1.1} & 95.3\dev{0.4} \\
Capsules & 90.9\dev{0.0} & 79.4\dev{4.9} & 63.4\dev{5.7} & 72.3\dev{5.3} & 89.9\dev{2.5} & 79.9\dev{5.8} & 63.4\dev{2.0} & 73.6\dev{4.7} & 88.9\dev{0.7} & 81.1\dev{4.5} & 63.0\dev{2.3} & 78.4\dev{3.1} & 91.5\dev{1.4} \\
Cashew & 96.4\dev{0.0} & 97.9\dev{0.4} & 78.2\dev{5.7} & 94.6\dev{2.0} & 97.2\dev{0.2} & 98.6\dev{0.6} & 86.1\dev{2.2} & 96.9\dev{0.3} & 97.3\dev{0.2} & 98.3\dev{0.6} & 88.4\dev{2.0} & 96.5\dev{0.7} & 97.7\dev{0.4} \\
Chewinggum & 98.6\dev{0.0} & 96.4\dev{0.9} & 79.8\dev{3.6} & 98.9\dev{0.1} & 99.0\dev{0.3} & 97.1\dev{0.4} & 89.5\dev{1.9} & 99.1\dev{0.2} & 98.9\dev{0.3} & 97.1\dev{0.6} & 88.5\dev{3.2} & 99.3\dev{0.1} & 99.0\dev{0.1} \\
Fryum & 90.1\dev{0.0} & 89.8\dev{1.8} & 74.5\dev{2.9} & 87.6\dev{2.4} & 94.7\dev{1.0} & 94.5\dev{2.3} & 81.0\dev{5.4} & 92.1\dev{1.3} & 95.8\dev{0.2} & 95.8\dev{1.0} & 81.5\dev{3.0} & 95.0\dev{0.6} & 96.0\dev{0.3} \\
Macaroni1 & 75.8\dev{0.0} & 61.9\dev{11.2} & 60.4\dev{2.9} & 67.8\dev{3.4} & 84.9\dev{1.2} & 64.5\dev{9.5} & 63.1\dev{4.3} & 74.9\dev{5.2} & 84.7\dev{1.5} & 60.2\dev{2.7} & 64.9\dev{2.1} & 82.1\dev{3.5} & 86.5\dev{0.6} \\
Macaroni2 & 60.3\dev{0.0} & 52.7\dev{4.2} & 51.7\dev{5.0} & 54.9\dev{3.2} & 68.4\dev{1.8} & 55.9\dev{3.1} & 52.7\dev{1.5} & 57.2\dev{2.6} & 70.4\dev{1.8} & 51.9\dev{2.3} & 54.9\dev{2.5} & 60.2\dev{3.0} & 69.6\dev{2.8} \\
PCB1  & 78.4\dev{0.0} & 84.9\dev{3.7} & 68.6\dev{2.4} & 72.1\dev{2.5} & 76.5\dev{19.0} & 83.8\dev{2.1} & 60.4\dev{7.7} & 72.6\dev{16.4} & 78.3\dev{4.3} & 83.2\dev{7.2} & 77.4\dev{2.9} & 81.0\dev{9.2} & 87.7\dev{1.7} \\
PCB2  & 49.2\dev{0.0} & 74.9\dev{2.9} & 63.3\dev{1.2} & 84.4\dev{0.4} & 64.9\dev{3.3} & 71.7\dev{6.6} & 68.9\dev{2.6} & 86.6\dev{1.1} & 65.8\dev{4.0} & 74.2\dev{5.0} & 75.0\dev{1.7} & 86.2\dev{1.0} & 71.3\dev{3.4} \\
PCB3  & 76.5\dev{0.0} & 75.5\dev{2.1} & 52.3\dev{10.8} & 84.6\dev{1.5} & 73.5\dev{1.6} & 78.3\dev{5.2} & 65.2\dev{3.8} & 86.1\dev{0.5} & 80.9\dev{1.6} & 81.0\dev{3.6} & 64.5\dev{2.4} & 88.3\dev{1.1} & 84.8\dev{1.8} \\
PCB4  & 77.7\dev{0.0} & 92.9\dev{1.6} & 74.7\dev{2.6} & 92.8\dev{3.1} & 78.5\dev{15.5} & 81.9\dev{11.2} & 67.6\dev{11.9} & 93.2\dev{3.4} & 72.5\dev{16.2} & 94.8\dev{2.9} & 84.0\dev{2.0} & 94.9\dev{1.2} & 85.6\dev{8.9} \\
Pipe fryum & 82.3\dev{0.0} & 88.3\dev{2.0} & 79.2\dev{1.5} & 95.4\dev{0.6} & 98.6\dev{0.5} & 88.1\dev{1.7} & 84.5\dev{1.7} & 96.8\dev{0.7} & 99.0\dev{0.3} & 88.8\dev{1.0} & 89.8\dev{1.7} & 98.3\dev{0.3} & 99.2\dev{0.2} \\
\cmidrule(r){1-1} \cmidrule(lr){2-2} \cmidrule(lr){3-6} \cmidrule(lr){7-10} \cmidrule(l){11-14}
Mean  & \textbf{81.2\dev{0.0}} & 82.0\dev{3.3} & 68.3\dev{4.0} & 82.8\dev{2.3} & \textbf{85.1\dev{4.0}} & 82.3\dev{4.3} & 71.6\dev{3.8} & 84.8\dev{3.2} & \textbf{85.8\dev{2.7}} & 83.4\dev{2.7} & 75.6\dev{2.2} & 87.5\dev{2.1} & \textbf{88.8\dev{1.8}} \\
\bottomrule
\end{tabular}%

  \end{adjustbox}
  \caption{Comparison of anomaly classification (AC) performance in terms of class-wise AUPR on VisA. We report the mean and standard deviation over 5 random seeds for each measurement.}
  \label{tab:visa/ac/aupr}
  \vspace{0.1in}
  \begin{adjustbox}{width=\linewidth}
\begin{tabular}{cccccccccccccc}
\toprule
VisA (AC) & $K=0$ & \multicolumn{4}{c}{$K=1$}     & \multicolumn{4}{c}{$K=2$}     & \multicolumn{4}{c}{$K=4$} \\
\cmidrule(lr){2-2} \cmidrule(lr){3-6} \cmidrule(lr){7-10} \cmidrule(l){11-14}
$F_1$-max & \textbf{WinCLIP} & SPADE & PaDiM & PatchCore & \textbf{WinCLIP+} & SPADE & PaDiM & PatchCore & \textbf{WinCLIP+} & SPADE & PaDiM & PatchCore & \textbf{WinCLIP+} \\
\cmidrule(r){1-1} \cmidrule(lr){2-2} \cmidrule(lr){3-6} \cmidrule(lr){7-10} \cmidrule(l){11-14}
Candle & 89.4\dev{0.0} & 80.4\dev{6.1} & 72.0\dev{2.0} & 79.5\dev{0.6} & 87.8\dev{1.2} & 85.5\dev{2.9} & 74.8\dev{2.9} & 78.7\dev{0.8} & 89.1\dev{1.3} & 86.9\dev{2.4} & 76.7\dev{1.3} & 80.5\dev{0.9} & 88.9\dev{1.0} \\
Capsules & 83.9\dev{0.0} & 81.0\dev{2.4} & 77.5\dev{0.8} & 77.9\dev{1.5} & 84.9\dev{2.0} & 80.4\dev{3.9} & 77.2\dev{0.4} & 77.2\dev{0.3} & 85.4\dev{0.6} & 79.5\dev{1.8} & 77.2\dev{0.3} & 77.3\dev{0.6} & 86.0\dev{0.9} \\
Cashew & 88.4\dev{0.0} & 94.8\dev{1.8} & 80.8\dev{0.7} & 89.6\dev{3.6} & 90.7\dev{0.7} & 95.5\dev{2.2} & 82.4\dev{1.2} & 92.3\dev{0.5} & 90.9\dev{0.7} & 95.7\dev{0.9} & 82.5\dev{1.2} & 91.1\dev{2.1} & 91.6\dev{1.3} \\
Chewinggum & 94.8\dev{0.0} & 89.7\dev{2.2} & 83.8\dev{2.0} & 95.9\dev{0.8} & 95.6\dev{0.9} & 90.5\dev{1.2} & 86.7\dev{0.8} & 97.0\dev{0.5} & 95.4\dev{0.6} & 91.3\dev{1.5} & 87.9\dev{0.8} & 97.4\dev{0.6} & 95.7\dev{0.5} \\
Fryum & 82.7\dev{0.0} & 85.3\dev{2.1} & 80.6\dev{0.8} & 82.9\dev{1.7} & 87.2\dev{1.4} & 90.9\dev{1.8} & 82.7\dev{2.1} & 84.7\dev{1.4} & 88.4\dev{0.6} & 91.9\dev{1.7} & 82.9\dev{1.8} & 86.6\dev{0.6} & 88.9\dev{0.8} \\
Macaroni1 & 74.2\dev{0.0} & 71.9\dev{2.0} & 69.2\dev{2.3} & 70.4\dev{1.9} & 76.2\dev{1.4} & 72.8\dev{3.1} & 68.8\dev{1.8} & 74.3\dev{2.1} & 76.7\dev{2.0} & 70.8\dev{1.2} & 70.0\dev{1.6} & 78.9\dev{1.4} & 78.2\dev{1.2} \\
Macaroni2 & 69.8\dev{0.0} & 68.1\dev{0.8} & 67.1\dev{0.2} & 67.6\dev{0.7} & 72.3\dev{1.1} & 68.2\dev{1.2} & 67.1\dev{0.4} & 67.6\dev{1.2} & 73.9\dev{0.9} & 67.9\dev{0.6} & 68.4\dev{1.0} & 68.8\dev{0.8} & 73.1\dev{1.6} \\
PCB1  & 71.0\dev{0.0} & 85.5\dev{0.2} & 80.3\dev{0.8} & 84.5\dev{0.4} & 81.3\dev{6.6} & 85.8\dev{0.2} & 71.4\dev{5.4} & 78.3\dev{6.7} & 73.2\dev{3.7} & 81.2\dev{6.4} & 83.1\dev{0.6} & 85.6\dev{1.8} & 83.1\dev{2.2} \\
PCB2  & 67.1\dev{0.0} & 70.9\dev{1.9} & 68.6\dev{1.4} & 75.9\dev{0.8} & 67.2\dev{0.3} & 71.5\dev{2.7} & 69.2\dev{0.6} & 78.1\dev{2.1} & 67.3\dev{0.3} & 71.1\dev{3.2} & 72.0\dev{2.3} & 79.2\dev{1.9} & 67.7\dev{0.6} \\
PCB3  & 71.0\dev{0.0} & 70.2\dev{1.4} & 67.7\dev{0.9} & 76.7\dev{2.5} & 73.5\dev{1.5} & 73.3\dev{3.6} & 69.9\dev{1.1} & 78.9\dev{1.0} & 73.9\dev{1.3} & 75.5\dev{3.3} & 69.0\dev{0.7} & 80.7\dev{0.5} & 77.0\dev{1.4} \\
PCB4  & 74.9\dev{0.0} & 87.5\dev{1.7} & 74.5\dev{2.2} & 90.4\dev{2.7} & 86.1\dev{2.1} & 83.1\dev{5.9} & 74.6\dev{3.9} & 91.3\dev{4.1} & 86.8\dev{3.8} & 90.6\dev{2.1} & 81.0\dev{1.6} & 92.2\dev{3.4} & 84.6\dev{7.0} \\
Pipe fryum & 80.7\dev{0.0} & 82.7\dev{0.7} & 81.2\dev{0.5} & 89.2\dev{2.4} & 94.4\dev{0.7} & 82.7\dev{1.4} & 83.1\dev{0.6} & 91.4\dev{1.1} & 95.4\dev{0.8} & 82.8\dev{0.6} & 85.4\dev{0.8} & 93.9\dev{1.2} & 95.6\dev{0.7} \\
\cmidrule(r){1-1} \cmidrule(lr){2-2} \cmidrule(lr){3-6} \cmidrule(lr){7-10} \cmidrule(l){11-14}
Mean  & \textbf{79.0\dev{0.0}} & 80.7\dev{1.9} & 75.3\dev{1.2} & 81.7\dev{1.6} & \textbf{83.1\dev{1.7}} & 81.7\dev{2.5} & 75.7\dev{1.8} & 82.5\dev{1.8} & \textbf{83.0\dev{1.4}} & 82.1\dev{2.1} & 78.0\dev{1.2} & 84.3\dev{1.3} & \textbf{84.2\dev{1.6}} \\
\bottomrule
\end{tabular}%

  \end{adjustbox}
  \caption{Comparison of anomaly classification (AC) performance in terms of class-wise $F_1$-max on VisA. We report the mean and standard deviation over 5 random seeds for each measurement.}
  \label{tab:visa/ac/f1}
\end{table*}

\begin{table*}[!ht]
  \centering
  \begin{adjustbox}{width=\linewidth}
\begin{tabular}{cccccccccccccc}
\toprule
VisA (AS) & $K=0$ & \multicolumn{4}{c}{$K=1$}     & \multicolumn{4}{c}{$K=2$}     & \multicolumn{4}{c}{$K=4$} \\
\cmidrule(lr){2-2} \cmidrule(lr){3-6} \cmidrule(lr){7-10} \cmidrule(l){11-14}
pAUROC & \textbf{WinCLIP} & SPADE & PaDiM & PatchCore & \textbf{WinCLIP+} & SPADE & PaDiM & PatchCore & \textbf{WinCLIP+} & SPADE & PaDiM & PatchCore & \textbf{WinCLIP+} \\
\cmidrule(r){1-1} \cmidrule(lr){2-2} \cmidrule(lr){3-6} \cmidrule(lr){7-10} \cmidrule(l){11-14}
Candle & 88.9\dev{0.0} & 97.9\dev{0.3} & 91.7\dev{2.2} & 97.2\dev{0.2} & 97.4\dev{0.2} & 98.1\dev{0.2} & 94.9\dev{0.8} & 97.7\dev{0.3} & 97.7\dev{0.1} & 98.2\dev{0.1} & 95.4\dev{0.2} & 97.9\dev{0.1} & 97.8\dev{0.2} \\
Capsules & 81.6\dev{0.0} & 95.5\dev{0.5} & 70.9\dev{1.1} & 93.2\dev{0.9} & 96.4\dev{0.6} & 96.5\dev{0.9} & 75.7\dev{1.7} & 94.0\dev{0.2} & 96.8\dev{0.3} & 97.7\dev{0.1} & 79.1\dev{0.7} & 94.8\dev{0.5} & 97.1\dev{0.2} \\
Cashew & 84.7\dev{0.0} & 95.9\dev{0.5} & 95.5\dev{0.6} & 98.1\dev{0.1} & 98.5\dev{0.2} & 95.9\dev{0.4} & 96.4\dev{0.4} & 98.2\dev{0.2} & 98.5\dev{0.1} & 95.9\dev{0.3} & 97.2\dev{0.3} & 98.3\dev{0.2} & 98.7\dev{0.0} \\
Chewinggum & 93.3\dev{0.0} & 96.0\dev{0.4} & 90.1\dev{0.4} & 96.9\dev{0.3} & 98.6\dev{0.1} & 96.0\dev{0.3} & 93.1\dev{0.7} & 96.6\dev{0.1} & 98.6\dev{0.1} & 95.7\dev{0.3} & 94.4\dev{0.5} & 96.8\dev{0.1} & 98.5\dev{0.1} \\
Fryum & 88.5\dev{0.0} & 93.5\dev{0.3} & 93.3\dev{0.6} & 93.3\dev{0.5} & 96.4\dev{0.3} & 93.9\dev{0.2} & 94.1\dev{0.6} & 94.0\dev{0.3} & 97.0\dev{0.2} & 94.4\dev{0.1} & 95.0\dev{0.4} & 94.2\dev{0.2} & 97.1\dev{0.1} \\
Macaroni1 & 70.9\dev{0.0} & 97.9\dev{0.2} & 89.4\dev{0.9} & 95.2\dev{0.4} & 96.4\dev{0.6} & 98.5\dev{0.2} & 91.7\dev{0.3} & 96.0\dev{1.3} & 96.5\dev{0.7} & 98.8\dev{0.1} & 93.5\dev{0.5} & 97.0\dev{0.3} & 97.0\dev{0.2} \\
Macaroni2 & 59.3\dev{0.0} & 94.1\dev{1.0} & 86.4\dev{1.1} & 89.1\dev{1.6} & 96.8\dev{0.4} & 95.2\dev{0.4} & 90.1\dev{0.8} & 90.2\dev{1.9} & 96.8\dev{0.6} & 96.4\dev{0.2} & 90.2\dev{0.3} & 93.9\dev{0.3} & 97.3\dev{0.3} \\
PCB1  & 61.2\dev{0.0} & 94.7\dev{0.4} & 89.9\dev{0.3} & 96.1\dev{1.5} & 96.6\dev{0.6} & 96.5\dev{1.5} & 90.6\dev{0.6} & 97.6\dev{0.9} & 97.0\dev{0.9} & 96.8\dev{1.5} & 93.2\dev{1.5} & 98.1\dev{1.0} & 98.1\dev{0.9} \\
PCB2  & 71.6\dev{0.0} & 95.1\dev{0.2} & 90.9\dev{1.4} & 95.4\dev{0.2} & 93.0\dev{0.4} & 95.7\dev{0.1} & 93.9\dev{0.9} & 96.0\dev{0.3} & 93.9\dev{0.2} & 96.3\dev{0.0} & 93.7\dev{1.0} & 96.6\dev{0.2} & 94.6\dev{0.4} \\
PCB3  & 85.3\dev{0.0} & 96.0\dev{0.1} & 93.9\dev{0.3} & 96.2\dev{0.3} & 94.3\dev{0.3} & 96.6\dev{0.1} & 95.1\dev{0.5} & 97.1\dev{0.1} & 95.1\dev{0.2} & 96.9\dev{0.0} & 95.7\dev{0.1} & 97.4\dev{0.2} & 95.8\dev{0.1} \\
PCB4  & 94.4\dev{0.0} & 92.0\dev{0.6} & 89.6\dev{0.6} & 95.6\dev{0.6} & 94.0\dev{0.9} & 92.8\dev{0.3} & 90.7\dev{0.9} & 96.2\dev{0.4} & 95.6\dev{0.3} & 94.1\dev{0.2} & 92.1\dev{0.5} & 97.0\dev{0.2} & 96.1\dev{0.3} \\
Pipe fryum & 75.4\dev{0.0} & 98.4\dev{0.2} & 97.2\dev{0.6} & 98.8\dev{0.2} & 98.3\dev{0.2} & 98.7\dev{0.1} & 98.1\dev{0.4} & 99.1\dev{0.1} & 98.5\dev{0.2} & 98.8\dev{0.0} & 98.5\dev{0.1} & 99.1\dev{0.0} & 98.7\dev{0.1} \\
\cmidrule(r){1-1} \cmidrule(lr){2-2} \cmidrule(lr){3-6} \cmidrule(lr){7-10} \cmidrule(l){11-14}
Mean  & \textbf{79.6\dev{0.0}} & 95.6\dev{0.4} & 89.9\dev{0.8} & 95.4\dev{0.6} & \textbf{96.4\dev{0.4}} & 96.2\dev{0.4} & 92.0\dev{0.7} & 96.1\dev{0.5} & \textbf{96.8\dev{0.3}} & 96.6\dev{0.3} & 93.2\dev{0.5} & 96.8\dev{0.3} & \textbf{97.2\dev{0.2}} \\
\bottomrule
\end{tabular}%

  \end{adjustbox}
  \caption{Comparison of anomaly segmentation (AS) performance in terms of class-wise pixel-AUROC on VisA. We report the mean and standard deviation over 5 random seeds for each measurement.}
  \label{tab:visa/as/proc}
  \vspace{0.1in}
  \begin{adjustbox}{width=\linewidth}
\begin{tabular}{cccccccccccccc}
\toprule
VisA (AS) & $K=0$ & \multicolumn{4}{c}{$K=1$}     & \multicolumn{4}{c}{$K=2$}     & \multicolumn{4}{c}{$K=4$} \\
\cmidrule(lr){2-2} \cmidrule(lr){3-6} \cmidrule(lr){7-10} \cmidrule(l){11-14}
PRO   & \textbf{WinCLIP} & SPADE & PaDiM & PatchCore & \textbf{WinCLIP+} & SPADE & PaDiM & PatchCore & \textbf{WinCLIP+} & SPADE & PaDiM & PatchCore & \textbf{WinCLIP+} \\
\cmidrule(r){1-1} \cmidrule(lr){2-2} \cmidrule(lr){3-6} \cmidrule(lr){7-10} \cmidrule(l){11-14}
Candle & 83.5\dev{0.0} & 95.6\dev{0.5} & 81.5\dev{5.3} & 92.6\dev{0.4} & 94.0\dev{0.4} & 95.6\dev{0.4} & 87.3\dev{1.2} & 93.4\dev{0.6} & 94.2\dev{0.2} & 95.7\dev{0.1} & 88.3\dev{0.7} & 94.1\dev{0.4} & 94.4\dev{0.2} \\
Capsules & 35.3\dev{0.0} & 83.1\dev{1.1} & 30.6\dev{1.1} & 66.6\dev{4.5} & 73.6\dev{3.5} & 85.4\dev{3.1} & 38.4\dev{3.7} & 67.9\dev{2.3} & 75.9\dev{1.9} & 89.0\dev{1.2} & 43.3\dev{2.0} & 69.0\dev{3.2} & 77.0\dev{1.4} \\
Cashew & 76.4\dev{0.0} & 89.8\dev{1.1} & 73.4\dev{2.1} & 90.8\dev{0.2} & 91.1\dev{0.8} & 90.4\dev{0.5} & 78.4\dev{2.7} & 91.4\dev{1.0} & 90.4\dev{0.6} & 90.4\dev{0.6} & 81.2\dev{2.8} & 92.1\dev{0.3} & 91.3\dev{0.9} \\
Chewinggum & 70.4\dev{0.0} & 73.9\dev{1.2} & 58.1\dev{0.6} & 78.2\dev{1.3} & 91.0\dev{0.5} & 73.8\dev{1.1} & 63.7\dev{2.4} & 78.0\dev{0.4} & 90.9\dev{0.7} & 72.7\dev{0.9} & 67.2\dev{1.8} & 79.3\dev{0.8} & 91.0\dev{0.4} \\
Fryum & 77.4\dev{0.0} & 83.7\dev{1.2} & 71.1\dev{1.6} & 78.7\dev{2.3} & 89.1\dev{1.0} & 84.5\dev{0.9} & 71.2\dev{0.8} & 81.4\dev{2.8} & 89.3\dev{0.2} & 86.2\dev{0.9} & 73.2\dev{1.3} & 81.0\dev{1.2} & 89.7\dev{0.5} \\
Macaroni1 & 34.3\dev{0.0} & 92.0\dev{0.6} & 62.2\dev{4.4} & 83.4\dev{1.3} & 84.6\dev{2.3} & 93.9\dev{0.8} & 71.8\dev{2.4} & 86.2\dev{4.6} & 85.2\dev{1.4} & 95.1\dev{0.4} & 76.6\dev{2.1} & 89.6\dev{0.7} & 86.8\dev{0.8} \\
Macaroni2 & 21.4\dev{0.0} & 80.0\dev{3.3} & 54.9\dev{3.6} & 66.0\dev{3.0} & 89.3\dev{2.4} & 81.7\dev{1.5} & 65.6\dev{3.4} & 67.2\dev{6.5} & 88.6\dev{1.7} & 86.0\dev{0.8} & 65.9\dev{1.5} & 78.3\dev{0.9} & 90.5\dev{1.3} \\
PCB1  & 26.3\dev{0.0} & 81.3\dev{5.7} & 63.9\dev{1.8} & 79.0\dev{10.7} & 82.5\dev{6.0} & 87.2\dev{2.3} & 68.4\dev{4.1} & 86.1\dev{1.7} & 83.8\dev{5.0} & 88.0\dev{2.7} & 70.2\dev{3.3} & 88.1\dev{2.6} & 87.9\dev{2.1} \\
PCB2  & 37.2\dev{0.0} & 83.7\dev{0.6} & 64.4\dev{3.8} & 80.9\dev{0.5} & 73.6\dev{1.5} & 85.5\dev{1.0} & 72.9\dev{3.4} & 82.9\dev{1.8} & 76.2\dev{0.9} & 87.0\dev{0.5} & 71.9\dev{2.6} & 83.7\dev{1.0} & 78.0\dev{1.3} \\
PCB3  & 56.1\dev{0.0} & 84.3\dev{1.0} & 69.0\dev{1.2} & 78.1\dev{2.0} & 79.5\dev{2.5} & 86.1\dev{0.6} & 74.0\dev{2.3} & 82.2\dev{1.1} & 82.3\dev{1.8} & 87.7\dev{0.6} & 77.2\dev{0.8} & 84.4\dev{1.9} & 84.2\dev{1.0} \\
PCB4  & 80.4\dev{0.0} & 66.9\dev{2.0} & 59.1\dev{1.8} & 77.9\dev{3.1} & 76.6\dev{4.1} & 69.3\dev{1.1} & 62.6\dev{3.6} & 79.5\dev{4.8} & 81.7\dev{1.2} & 74.7\dev{1.0} & 67.9\dev{2.6} & 83.5\dev{2.5} & 84.2\dev{0.7} \\
Pipe fryum & 82.3\dev{0.0} & 94.3\dev{0.5} & 83.9\dev{0.8} & 93.6\dev{0.5} & 96.1\dev{0.6} & 95.0\dev{0.2} & 86.9\dev{0.9} & 94.5\dev{0.4} & 96.2\dev{0.6} & 95.0\dev{0.3} & 88.7\dev{1.3} & 95.0\dev{0.5} & 96.6\dev{0.2} \\
\cmidrule(r){1-1} \cmidrule(lr){2-2} \cmidrule(lr){3-6} \cmidrule(lr){7-10} \cmidrule(l){11-14}
Mean  & \textbf{56.8\dev{0.0}} & 84.1\dev{1.6} & 64.3\dev{2.4} & 80.5\dev{2.5} & \textbf{85.1\dev{2.1}} & 85.7\dev{1.1} & 70.1\dev{2.6} & 82.6\dev{2.3} & \textbf{86.2\dev{1.4}} & 87.3\dev{0.8} & 72.6\dev{1.9} & 84.9\dev{1.4} & \textbf{87.6\dev{0.9}} \\
\bottomrule
\end{tabular}%

  \end{adjustbox}
  \caption{Comparison of anomaly segmentation (AS) performance in terms of class-wise PRO on VisA. We report the mean and standard deviation over 5 random seeds for each measurement.}
  \label{tab:visa/as/pro}
  \vspace{0.1in}
  \begin{adjustbox}{width=\linewidth}
\begin{tabular}{cccccccccccccc}
\toprule
VisA (AS) & $K=0$ & \multicolumn{4}{c}{$K=1$}     & \multicolumn{4}{c}{$K=2$}     & \multicolumn{4}{c}{$K=4$} \\
\cmidrule(lr){2-2} \cmidrule(lr){3-6} \cmidrule(lr){7-10} \cmidrule(l){11-14}
$F_1$-max & \textbf{WinCLIP} & SPADE & PaDiM & PatchCore & \textbf{WinCLIP+} & SPADE & PaDiM & PatchCore & \textbf{WinCLIP+} & SPADE & PaDiM & PatchCore & \textbf{WinCLIP+} \\
\cmidrule(r){1-1} \cmidrule(lr){2-2} \cmidrule(lr){3-6} \cmidrule(lr){7-10} \cmidrule(l){11-14}
Candle & 22.5\dev{0.0} & 37.1\dev{0.8} & 19.6\dev{2.6} & 40.9\dev{1.0} & 42.7\dev{1.7} & 37.6\dev{0.5} & 21.6\dev{1.4} & 40.4\dev{1.0} & 42.2\dev{0.8} & 38.3\dev{1.0} & 21.6\dev{0.7} & 41.0\dev{1.0} & 43.0\dev{0.9} \\
Capsules & 9.2\dev{0.0} & 28.5\dev{7.8} & 2.3\dev{0.2} & 35.7\dev{6.8} & 58.2\dev{1.3} & 37.4\dev{8.4} & 3.0\dev{0.4} & 37.8\dev{5.7} & 57.0\dev{3.7} & 48.0\dev{2.0} & 3.9\dev{0.4} & 47.0\dev{3.0} & 59.8\dev{1.8} \\
Cashew & 13.2\dev{0.0} & 51.1\dev{1.4} & 35.1\dev{4.9} & 60.4\dev{1.1} & 59.5\dev{2.1} & 52.8\dev{1.2} & 40.8\dev{4.0} & 60.3\dev{0.4} & 60.5\dev{2.4} & 54.1\dev{0.5} & 47.2\dev{2.9} & 60.7\dev{0.4} & 62.3\dev{1.1} \\
Chewinggum & 41.1\dev{0.0} & 58.7\dev{1.1} & 19.7\dev{2.0} & 64.5\dev{1.0} & 65.3\dev{0.5} & 59.9\dev{0.5} & 29.5\dev{6.4} & 63.9\dev{0.4} & 64.8\dev{0.9} & 59.5\dev{0.6} & 37.8\dev{4.4} & 64.4\dev{0.6} & 65.2\dev{0.2} \\
Fryum & 22.1\dev{0.0} & 34.0\dev{1.5} & 32.3\dev{1.5} & 37.2\dev{1.4} & 50.8\dev{1.8} & 36.6\dev{2.1} & 36.5\dev{3.6} & 41.1\dev{2.9} & 54.8\dev{1.7} & 40.3\dev{1.8} & 44.5\dev{1.3} & 44.6\dev{2.9} & 56.5\dev{0.6} \\
Macaroni1 & 7.0\dev{0.0} & 28.5\dev{3.4} & 4.8\dev{0.7} & 16.5\dev{2.6} & 34.1\dev{1.7} & 39.2\dev{3.5} & 5.5\dev{1.2} & 20.0\dev{8.2} & 33.2\dev{1.9} & 37.6\dev{5.8} & 7.0\dev{0.7} & 21.6\dev{1.9} & 33.8\dev{0.9} \\
Macaroni2 & 1.0\dev{0.0} & 6.7\dev{2.8} & 1.9\dev{0.5} & 2.7\dev{0.7} & 34.4\dev{3.0} & 8.6\dev{1.8} & 2.4\dev{0.1} & 5.1\dev{2.3} & 29.9\dev{3.4} & 18.3\dev{3.0} & 2.4\dev{0.3} & 10.9\dev{1.7} & 35.1\dev{2.5} \\
PCB1  & 2.4\dev{0.0} & 16.6\dev{1.1} & 8.6\dev{0.7} & 30.7\dev{3.2} & 25.9\dev{2.6} & 34.8\dev{19.9} & 8.9\dev{0.5} & 47.7\dev{19.7} & 34.6\dev{16.2} & 37.1\dev{20.8} & 13.9\dev{3.6} & 55.9\dev{20.3} & 50.9\dev{20.4} \\
PCB2  & 4.7\dev{0.0} & 35.0\dev{1.9} & 9.8\dev{2.2} & 33.3\dev{1.0} & 18.7\dev{1.5} & 39.2\dev{1.1} & 16.6\dev{2.5} & 33.3\dev{0.6} & 24.0\dev{1.1} & 42.6\dev{1.0} & 15.8\dev{3.0} & 33.6\dev{0.3} & 27.8\dev{1.9} \\
PCB3  & 10.3\dev{0.0} & 43.9\dev{1.6} & 18.1\dev{0.3} & 36.6\dev{0.6} & 31.2\dev{6.7} & 44.9\dev{0.7} & 20.6\dev{1.6} & 37.1\dev{0.2} & 37.1\dev{2.8} & 47.5\dev{0.7} & 22.4\dev{1.3} & 37.3\dev{0.1} & 42.5\dev{1.1} \\
PCB4  & 32.0\dev{0.0} & 30.7\dev{1.0} & 13.3\dev{1.5} & 34.8\dev{1.4} & 22.8\dev{2.9} & 35.0\dev{3.4} & 15.5\dev{1.6} & 40.9\dev{4.3} & 30.6\dev{2.4} & 39.2\dev{5.2} & 19.8\dev{2.5} & 44.1\dev{4.4} & 31.9\dev{3.0} \\
Pipe fryum & 12.3\dev{0.0} & 55.5\dev{1.9} & 43.3\dev{3.6} & 62.6\dev{2.6} & 51.8\dev{2.0} & 59.3\dev{1.4} & 52.9\dev{5.5} & 64.5\dev{0.6} & 53.6\dev{2.6} & 61.1\dev{0.7} & 58.2\dev{0.8} & 65.0\dev{0.5} & 55.1\dev{1.1} \\
\cmidrule(r){1-1} \cmidrule(lr){2-2} \cmidrule(lr){3-6} \cmidrule(lr){7-10} \cmidrule(l){11-14}
Mean  & \textbf{14.8\dev{0.0}} & 35.5\dev{2.2} & 17.4\dev{1.7} & 38.0\dev{1.9} & \textbf{41.3\dev{2.3}} & 40.5\dev{3.7} & 21.1\dev{2.4} & 41.0\dev{3.9} & \textbf{43.5\dev{3.3}} & 43.6\dev{3.6} & 24.6\dev{1.8} & 43.9\dev{3.1} & \textbf{47.0\dev{3.0}} \\
\bottomrule
\end{tabular}%

  \end{adjustbox}
  \caption{Comparison of anomaly segmentation (AS) performance in terms of class-wise $F_1$-max on VisA. We report the mean and standard deviation over 5 random seeds for each measurement.}
  \label{tab:visa/as/pf1}
\end{table*}

\end{document}